\documentclass[11pt]{article} %11 is 11 pt font, can be 10, 11, or 12.
\usepackage[top=1in,bottom=1in,left=1in,right=1in]{geometry}
\usepackage{graphicx} %new package
\usepackage{setspace}
\usepackage{lscape}
\usepackage{natbib}
\usepackage{xcolor} %to make text in color
\usepackage{amsmath, amsfonts, amsthm}
\usepackage{microtype}

\usepackage{tabularx}

\usepackage{fancyhdr} 
\usepackage{lscape}
\usepackage{graphicx}  
\usepackage[update,prepend]{epstopdf} 
\usepackage{wrapfig}
\usepackage{multicol}
\usepackage{float}
\usepackage{multirow}
\usepackage{chngcntr}
\usepackage{pdfpages}
\usepackage{caption}
\usepackage{palatino}
\usepackage{booktabs}
\usepackage[font=bf]{caption} 
\usepackage{rotating}
\usepackage[font=footnotesize]{subcaption}
\usepackage{natbib}
\usepackage[colorlinks, citecolor=purple]{hyperref}

\definecolor{myLightGray}{RGB}{191,191,191}
\definecolor{myGray}{RGB}{160,160,160}
\definecolor{myDarkGray}{RGB}{144,144,144}
\definecolor{myDarkRed}{RGB}{167,114,115}
\definecolor{myRed}{RGB}{255,58,70}
\definecolor{myGreen}{RGB}{0,255,71}

\usepackage{geometry}
\usepackage{caption, subcaption}
\usepackage{pdflscape}

\usepackage[font=bf]{caption} %makes bold captions
\usepackage{rotating}
\usepackage[font=footnotesize]{subcaption}
\usepackage{tikz}
\usetikzlibrary{decorations.pathreplacing,calligraphy}
\usepackage{standalone}

\title{Decoding Visual Sentiment of Political Imagery}

\author{Olga Gasparyan\footnote{The authors are listed in alphabetical order with equal contributions to the project. All errors are ours.} \\ Florida State University \\ \href{mailto:ogasparyan@fsu.edu}{ogasparyan@fsu.edu}\and Elena Sirotkina\footnotemark[\value{footnote}] \\ UNC at Chapel Hill \\ \href{mailto:esirotkina@unc.edu}{esirotkina@unc.edu}}

\date{\today}

\begin{document}
\maketitle
\begin{abstract}
\singlespacing \noindent
How can we define visual sentiment when viewers systematically disagree on their perspectives? This study introduces a novel approach to visual sentiment analysis by integrating attitudinal differences into visual sentiment classification. Recognizing that societal divides, such as partisan differences, heavily influence sentiment labeling, we developed a dataset that reflects these divides. We then trained a deep learning multi-task multi-class model to predict visual sentiment from different ideological viewpoints. Applied to immigration-related images, our approach captures perspectives from both Democrats and Republicans. By incorporating diverse perspectives into the labeling and model training process, our strategy addresses the limitation of label ambiguity and demonstrates improved accuracy in visual sentiment predictions. Overall, our study advocates for a paradigm shift in decoding visual sentiment toward creating classifiers that more accurately reflect the sentiments generated by humans.

\end{abstract}

\newpage
\doublespacing
\section*{Introduction}
The rapid advance of visual social media platforms like Instagram, TikTok, and X\footnote{Formerly known as Twitter}, confirms our shift towards a more visual-centric world where images crucially direct our attention (see, e.g., \citealt{domke2002primes, grabe2014image, schill2012visual, chavez2023covering}), shape attitudes and sentiments \citep{bossetta2023cross, casas2019images}, and reinforce stereotypes \citep{carpinella2021visual, domke2002primes}. This effect is especially strong in politics, where the way events are visually depicted can influence public reaction \citep{grabe2009image}, particularly during pivotal times like elections \citep{grabe2014image} or when tensions around certain political issues rise \citep{de2023visual}.

\begin{figure}[tbh!]
\vspace{5mm}
    \centering
    \caption{How Would You Label Visual Sentiment of These Images?}
    \label{images12}
    \begin{minipage}{0.46\textwidth}
        \centering
         \caption*{\textbf{Image A}}
        \label{1a}
        \includegraphics[width=0.9\textwidth]{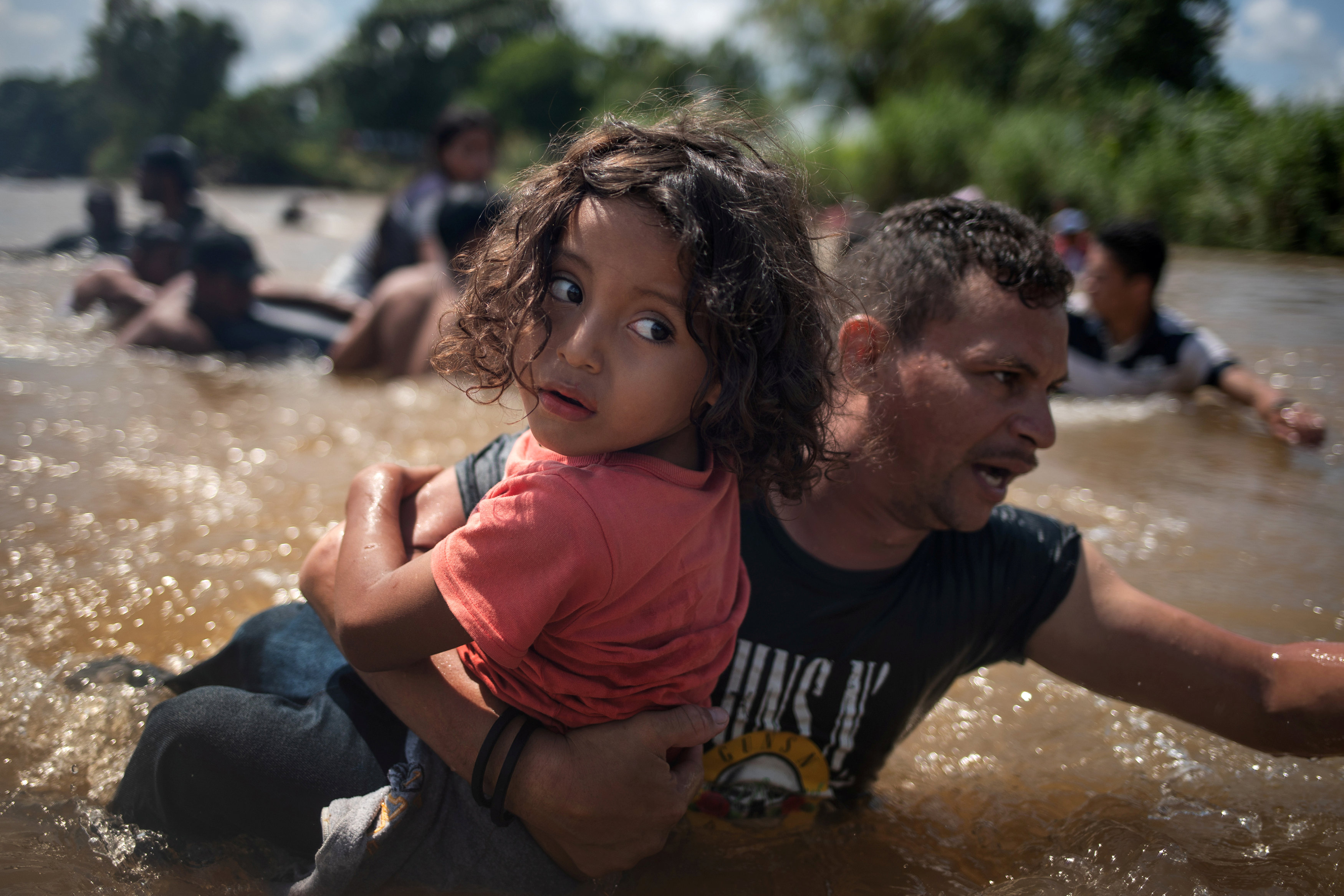}
        \footnotesize Published by Reuters (\hyperlink{https://widerimage.reuters.com/story/central-american-migrants-trek-north-to-seek-a-better-life}{Oct 29, 2018}). Photo by Reuters/Adrees Latif.
    \end{minipage}
    \hfill
    \begin{minipage}{0.46\textwidth}
        \centering
        \caption*{\textbf{Image B}}
        \label{1b}
        \includegraphics[width=\textwidth]{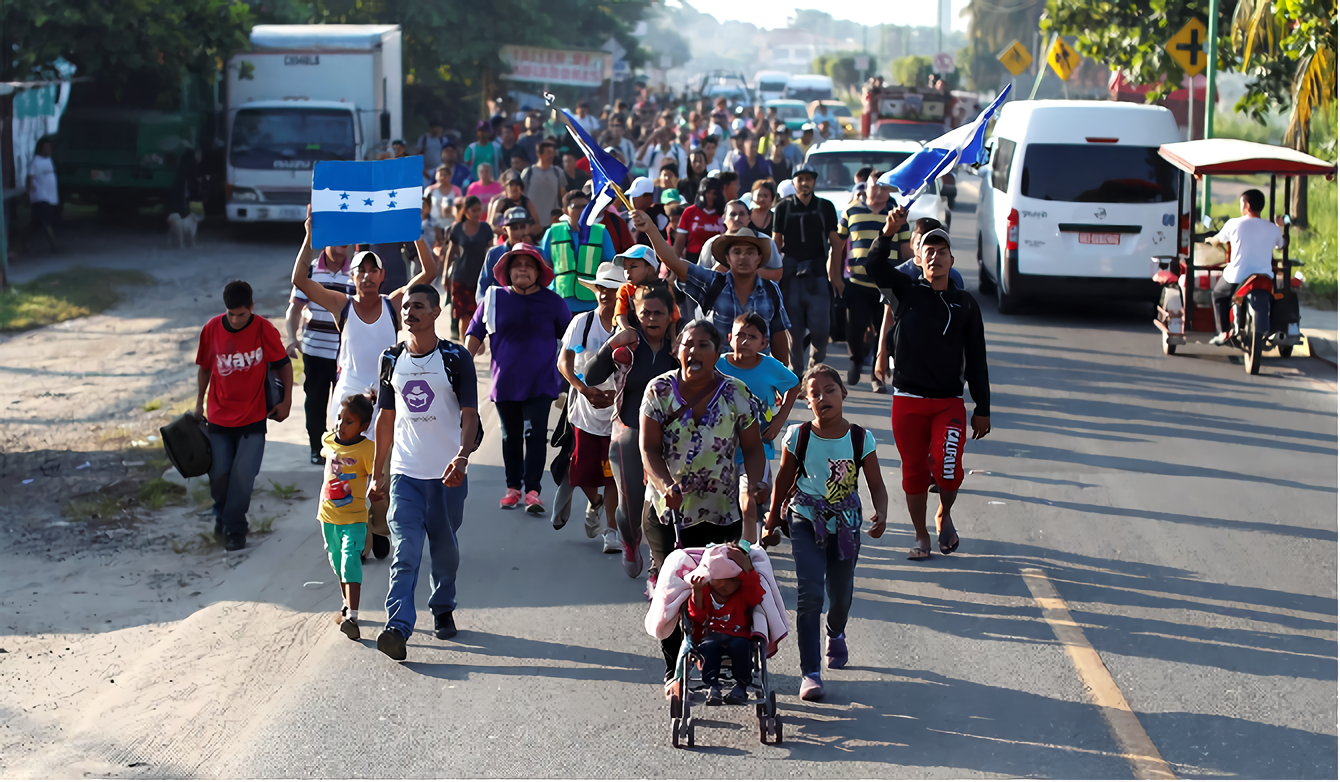}
        \footnotesize Published by National Review (\hyperlink{https://www.nationalreview.com/2018/10/migrant-caravan-democrats-border-enforcement-immigration-debate/}{Oct 26, 2018}). Photo by Reuters/Carlos Garcia Rawlins.
    \end{minipage}
\vspace{5mm}    
\end{figure}

To illustrate the latter point, consider images A and B in Figure \ref{images12}. Image A captures a young girl with her likely father as they cross a perilous river, while Image B depicts a crowd, resembling  protesters, with a discernible resolve. Both images, sourced from official US media accounts on X, accompany news stories about migrant caravans traveling from Central America. How should we approach these visuals to \textit{accurately} interpret their visual sentiment?

Computer vision research delineates the nature of visual sentiment through two primary content-focused approaches. One approach treats images holistically (\citealt{ortis2020survey}), compiling them into visual databanks (\citealt{peng2016emotions, kossaifi2019sewa}) that feature a diverse set of images tagged with sentiment labels (\citealt{washington2021training}) and enriched with metadata such as objects and scenes (\citealt{kim2023d}). These databanks facilitate the training of visual sentiment classifiers.

The other, object-centric approach focuses on "affective regions" within images (\citealt{yang2018visual, zhao2021affective}). These regions, established using techniques like Bag of Visual Words (BoVW) \citep{yang2010bag}, are considered more powerful at evoking certain attitudes than other regions \citep{zhao2021affective}. For example, a person's smile in an image may be tagged as an affective region that elicits a "positive" sentiment (\citealt{borth2013large}), whereas the presence of firearms tends to be labeled with "negative" sentiment (\citealt{chandrasekaran2022visual, you2016building}). These affective regions are then consolidated into sentiment maps that depict the overall emotional tone of the image (\citealt{chen2014deepsentibank}).

Crucially, the leading content-focused approaches in visual sentiment analysis assume that the sentiment conveyed by an image is an inherent attribute of the image itself \citep{chen2014deepsentibank, joo2018image, lan2023image}. This assumption leads these models to optimize computational techniques that focus on object and region identifications \citep{you2015robust, song2018boosting, wu2020visual}, as well as to improve sentiment dictionaries (see e.g. \citealt{you2016building}). However, such approaches seem to make a fundamental mistake. They overlook the critical importance of subjective personal attitudes and contextual factors on the sentiments that images generate \citep{webbwilliams2023}.

Recall that computational visual sentiment analysis was originally designed to closely mimic how humans attach sentiments and attitudes to images \citep{pang2015deep, fan2018emotional, wang2023unlocking}. Despite this, modern computer vision techniques often underplay the importance of subjective interpretations \citep{washington2021training}. Predicting the sentiment of an image, in contrast, should heavily depend on how viewers interpret symbols, colors, and compositions \citep{bar1996spatial, bar2004visual}, as well as their familiarity with and attitudes toward cultural, social, and political contexts depicted \citep{oosterhoff2018disgust}.

Now consider examples A and B again. The subjective sentiment that human coders assign to these images should largely depend on their attitudes toward the political issue depicted. It is empirically unlikely that a person (a coder) with a strong negative bias against immigration to the US will have the same sentiment about these images as someone who sympathizes with immigrants.

To address the current methodological limitations, we propose a novel approach in this study. When training a visual sentiment classifier for political images, we incorporate important attitudinal differences that people might have about the depicted topic. Hence, we propose the following workflow:

\begin{enumerate}
\item[] \textbf{(1) Identifying an Attitudinal Cleavage:} \
First, we examine whether visual labeling reflects a stable societal gap, such as a political divide in the USA, which must be embedded in the model training. However, getting there is a challenge. Creating one label that averages sentiments for particular visuals would result in a “neutral” label for these images, canceling out the sentiments of strong pro-issue and against-issue people among coders. This approach will fail to capture the \textit{true} sentiments that people actually hold.\footnote{Labeling at scale, as proposed by \citeauthor{benoit2016crowd} (2016), is considered one of the most promising solutions to this problem, at least with text; however, this approach may exacerbate the issue by averaging out polarized sentiments, thus masking the true diversity of opinions.} Using separate models for different attitude groups can lead to fragmented analysis and fail to capture the informative contradictory nature of visual sentiment in an image. Additionally, this approach may complicate the interpretation of results.

To address these limitations, we propose a single multi-task multi-class classification model that predicts multiple sentiment labels for disagreeing groups of coders. This approach helps mitigate the potential bias arising from individuals with different attitudinal priors assigning opposing sentiment labels.

\item[] \textbf{(2) Creating a Dataset of Sentiment Labels:} \
We constructed our own dataset of sentiment labels for a set of images about immigration, a politically polarizing topic where Democrats and Republicans typically disagree, representing a stable cleavage in political attitudes. We conducted two survey waves on a US sample, collecting coders’ socio-demographics and attitudes for 816 images depicting immigration. We then aggregated individual attitude scores at the image level and created partisan-based pairs of labels for sentiment outcomes\footnote{We asked participants to identify themselves as Democrats, Republicans, Independent or Other, and used only the responses from Democrats and Republicans for further analysis.} that reflect these attitudinal cleavages.

Importantly, what is known as \textit{'sentiment'} in computational social sciences corresponds to multiple related constructs in political and social psychology. We demonstrate how a nuanced approach to sentiment identification can enrich sentiment analysis by revealing varying degrees of perceptual divides. We empirically identify and discuss instances where these sentiments influence, and do not influence, the accuracy of visual sentiment labeling.

\item[] \textbf{(3) Training a Multi-task Multi-class Classifier:} \
We then use these constructed label pairs (corresponding to coder groups of Democrats and Republicans) to build visual sentiment classification models with a transfer learning approach. We adapt and fine-tune well-known very deep neural networks (ResNet50V2, DenseNet-121, and DenseNet-169) using our labeled data. These networks are designed to perform two tasks: (1) multi-task multi-class classification, categorizing sentiment as negative, neutral, or positive for two partisan groups (Democrats and Republicans), and (2) multi-task linear prediction, rating sentiment on a continuous scale [1,7] for the same groups. We ensure the robustness of our models by using K-fold cross-validation, which helps verify their accuracy and reliability.

\item[] \textbf{(4) Verification:} \
We use test examples of images with polarizing sentiment labels\footnote{For example, an image might be labeled as 'positive' by Democrats and 'negative' by Republicans.} to evaluate our classification model, which provides separate sentiment labels for Democrats and Republicans.
\end{enumerate}

Broadly, our study and findings advocate for a shift in computer vision regarding how we approach training classifiers for predicting sentiment and other attitudinal outcomes at scale. Instead of relying on a single, uniform label for each image, our findings argue for the importance of accounting for diverse perspectives, particularly those stemming from partisan and ideological differences, especially if political cleavages are stable and definitive in explaining how people feel about political images.

From a practical perspective, if certain images depicting polarizing topics systematically appear on social media, it is valuable to understand how they are received by partisans with different ideological positions. Our classifier helps quickly determine whether these images elicit similar associations or potentially exacerbate partisan divides. Our approach provides a straightforward, hands-on method for gaining this information.

Importantly, we caution that while partisanship is a significant cleavage for certain political issues and contexts—such as the U.S. with its strong and durable political polarization—other sensitive topics, both within and outside of politics, require consideration of different meaningful societal splits that shape people’s reactions to sensitive visuals. This necessitates a robust theoretical understanding of the relevant cleavages. However, without incorporating this information into the training process, attempts to predict "as-if" human sentiments without understanding how these sentiments actually form may be futile.
 
\section*{Visual Sentiment Labeling} 

\subsubsection*{Data Acquisition}
Since we are interested in political imagery and anticipate a divide in sentiment perceptions along ideological lines, we have chosen immigration as a political issue that systematically polarizes attitudes between Democrats and Republicans. 

Unlike other polarizing topics such as gun control, which often involve explicit affective objects (e.g., guns), immigration as a visual subject does not have universally recognizable visual cues that strongly signal emotional reactions. This makes it particularly valuable for examining how visuals can elicit diverse sentiments based solely on contextual understanding and symbolic elements rather than overtly affective objects.

We collected images covering real-live events accessed through publicly available social media accounts as well as stock image sources (such as Getty) that often serve as a primary source of imagery for many media outlets covering news on immigration. In total, we collected 816 images: 315 images are coming from the X (formerly Twitter) of U.S. media outlets\footnote{We used official X (formerly Twitter) handles of 393 U.S. media outlets to query all tweets containing the term "migrant caravan" between December 2017 and October 2021. Focusing on tweets with images, we created a subset of over 2,000 tweets (see Gasparyan and Sirotkina, \citeyear{gasparyan2023} for a detailed description of the dataset).} and 501 images with the same migrant caravan descriptions harvested from stock image sources.

\subsubsection*{Data Labeling Task}

How to define a sentiment as a function of human perception? ‘Visual sentiment’ in computer vision differs from the concept of sentiment in political and social psychology, which broadly refers to a viewer’s subjective perception of (an image's) tone and emotional message (see e.g., \citealt{smith1989sentiment}). To address this, we conducted two surveys\footnote{Both surveys were conducted on the Lucid Theorem platform: the first with 3,000 participants in April 2022 and the second with 2,000 participants in March 2024.} where respondents answered standard socio-demographic questions, self-identified as either Democrats or Republicans\footnote{Respondents who identified as Independents or from other parties were excluded.}, and evaluated ten randomly presented images based on their perceptions of the images’ tone and the sentiments they evoke, using the following variables:

\begin{enumerate}

\item \textit{"Sentiment"}: "Would you say that this image portrays the subject(s) or objects(s) in this picture in a positive or negative light? '1' stands for negative, '4' is neutral, and '7' stands for positive."

This is one of the most generic ways to measure sentiment (see e.g., \citealt{tang2009survey}). Previous literature shows that images of women and children tend to evoke more sympathetic attitudes \citep{carpenter2005women, bauer2018visual}, while images of protesting or transgressing crowds of men are more likely to be perceived negatively \citep{batziou2011framing,olier2022stereotypes}. This suggests that people's attitudes about images will be influenced both by how immigration is portrayed and by their attitudes toward immigration.

People with positive attitudes about immigration, such as many Democrats, are more likely to perceive women and children in a positive light. Conversely, Republicans, who often hold more critical views on immigration, are likely to view images of protesting men more negatively. This difference in perception is expected to produce a consistent partisan gap in the sentiment labels assigned to respective images.

\item \textit{"Subject of harm"}: "In your opinion, the subject(s) who is (are) portrayed in this picture is (are) more likely to be dangerous or harmless? '1' stands for dangerous, '4' is the middle ground, and '7' stands for harmless."

Republicans should be generally more likely to perceive certain immigrants as a potential threat, especially if the visual frame aligns with this perception. For instance, Republicans should be more likely to perceive a crowd of men as dangerous compared to women with children. In contrast, Democrats will view the same subjects as less threatening and more harmless, which aligns with their more lenient views on immigration and social inclusion. The perceived threat from the subjects in immigration-related images is an important component of the overall sentiment people attach to these images \citep{madrigal2023migrants}. By using the 'Subject of Harm' variable, we can effectively measure this aspect of visual sentiment that is relevant specifically to this topic.

\item \textit{"Object of harm"}: "In your opinion, the subject(s) who is (are) portrayed in this picture is (are) more likely to be vulnerable or safe? '1' stands for vulnerable, '4' is the middle ground, and '7' stands for safe."

Typically, Democrats may view immigrants with greater empathy, often perceiving subjects, especially women and children, as more vulnerable.  In contrast, Republicans may adopt a more critical stance on immigration, potentially viewing the same subjects as being safe and secure.

The 'Subjects of Harm' and 'Objects of Harm' variables represent two key dimensions in the partisan debate on immigration entailing that these variables are critical in defining the sentiments that Democrats and Republicans experience in response to various visual representations of immigration.

\item \textit{"Accuracy"}: "Do you think that this image is a faulty or accurate representation of the story that actually occurred? '1' stands for faulty, '4' is the middle ground, '7' stands for accurate."

The perception of accuracy is tied to the level of trust in the image as a source of information about the event that occurred \citep{kohring2007trust}. Partisans are expected to judge the accuracy of an image based on their perception of norms regarding how immigration issues should be depicted to illustrate the topic accurately \citep{fahmy2006visual}. People with different partisan perspectives might find images more trustworthy if those images represent immigration in a way that aligns with their pre-existing beliefs and stereotypes. Therefore, Democrats and Republicans may rate the accuracy of the same image differently, based on whether the image's tone and content match their individual perspectives.

\end{enumerate}

This approach to sentiment labeling is unconventional in computational literature, where labels are typically generated either through large-scale coding via crowdsourcing platforms or through expert-level coding by a few research assistants (see e.g. \citealt{rudinac2013learning}). In contrast, our approach attempts to illustrate how the individual characteristics of coders play a crucial role in shaping the sentiment labels.

\subsubsection*{Cleavages in Labeling and Image-Level Labels}
In complex tasks like assigning sentiment to images of politically sensitive topics, labels can reflect broader social and political disagreements influenced by fundamental factors such as gender, age, or race \citep{webbwilliams2023}. In our study, we concentrate on a politically polarizing issue, anticipating that visual sentiment assignments will vary across party lines. Therefore, we first generate image-level labels and then analyze whether there are significant differences in sentiment labeling between partisan groups.

We use respondents' partisanship, coded as Democrats (1) and Republicans (0)\footnote{In the survey, respondents self-identified as either Democrats or Republicans, and we excluded all others.}, to construct labels for each image. We then aggregate the evaluations of each image by calculating the mean score from all respondents. As a result, each image is assigned three scores: 1) the overall average score across all participants; 2) the average score from Democratic participants; and 3) the average score from Republican participants. This process is repeated for four outcomes of interest—sentiment, accuracy, subject of harm, and object of harm—to measure visual sentiment.

\begin{figure}[tbh!]
\vspace{1cm}
    \caption{Distribution of Image Evaluation Scores by Party}
    \label{dist_party}
    \centering
    \includegraphics[width=1\linewidth]{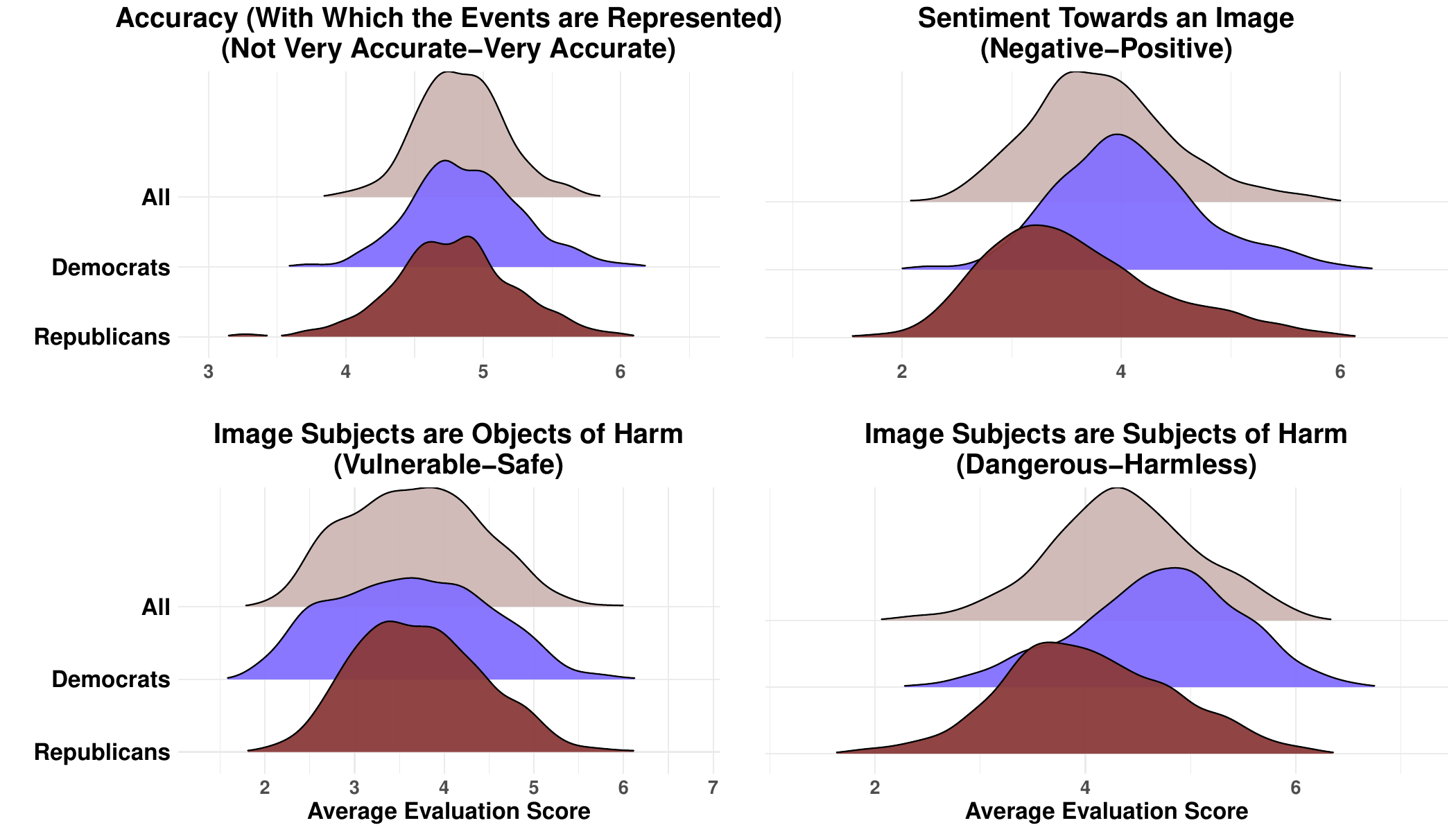}
    \begin{minipage}{16.5cm}
        \footnotesize{\textit{Note:} The plots show the density distribution of average evaluation scores across 816 images. The grey-shaded plots represent the average scores calculated for each image across all respondents. The blue-shaded plot shows the distribution of average evaluation scores for each image based solely on evaluations from respondents who self-identified as Democrats. The red-shaded density plot shows the distribution of average evaluation scores for each image based on evaluations from respondents who self-identified as Republicans.}
    \end{minipage}
\vspace{1cm}
\end{figure}

Figure \ref{dist_party} displays density plots of average evaluation scores for four sentiment measures across Democrats (blue-shaded plot), Republicans (red-shaded plot), and the overall average (gray-shaded plot). The plots indicate that Democrats and Republicans provide similar evaluations regarding image accuracy and the depiction of people as objects of harm, suggesting that averaging scores across all respondents does not introduce significant bias for these measures. However, for the 'Sentiment' and 'Subjects of Harm' variables, there are notable differences in average scores between Democrats and Republicans. This divergence implies that using a single label for these variables could introduce significant bias when labeling is performed at scale. Therefore, it is more appropriate to use separate labels for Democrats and Republicans for these variables in each image.

\subsubsection*{Label Classes}
Since we average respondents' evaluations, our image-level labels are measured on an interval scale [1,7]. While we use these interval scale labels for linear prediction tasks in our analysis, we also categorize them into a more conventional sentiment categorization. Specifically, we divide the interval scores into three categories: negative, neutral, and positive sentiments for both 'Sentiment' and 'Subject of Harm'\footnote{Since for the variable 'Subject of Harm' was originally measured as a range from dangerous to harmless, we proxy that the more dangerous the image subjects are perceived the more negative sentiment is assumed, conversely the more harmless the image subjects are perceived the more positive sentiment is assumed.} proxies of visual sentiment. To assign these categorical labels to the interval scale average evaluation scores (AES), we partition the range $\text{AES} \in [1,7]$ into three segments:
$$\text{Categorical Label}
=
\left \lbrace
\begin{array}{rcl}
\text{negative}, \hspace{0,2cm}  \text{if AES} \leq 3 \\
\text{neutral}, \hspace{0,2cm} 3 <  \text{if AES} < 5 \\
\text{positive}, \hspace{0,2cm}  \text{if AES} \geq 5 \\
\end{array} \right.$$

Table \ref{labels_dist} presents the distribution of images categorized by sentiment labels for Democrats and Republicans. The table shows that most images are labeled with a neutral sentiment. However, images rated by Democrats tend to receive more positive sentiment labels compared to those rated by Republicans. This difference points to the sentiment heterogeneity across partisan groups, which we should account for in visual sentiment prediction models. 

\begin{table}[!htbp]
\vspace{5mm}    
\caption{Distribution of Labeled Images Across Sentiment Categories}
\label{labels_dist}
\centering
\begin{tabular}{r|c|c|c|c}
  \hline
 &\multicolumn{2}{c}{Sentiment} & \multicolumn{2}{|c}{Subject of Harm}\\ \hline
 &  Democrats &  Republicans & Democrats &  Republicans \\ \hline
  \hline
Negative &  32 & 212 &  20 &  77 \\ 
Neutral & 701 & 547 & 493 & 630 \\ 
Positive &  83 &  57 & 303 & 109 \\ 
   \hline
\end{tabular}
\vspace{5mm}    
\end{table}

\section*{Visual Sentiment Prediction With Deep Learning Approach}

\subsubsection*{Transfer Learning From Pre-Trained Deep CNNs}
To predict visual sentiments, we adopt a supervised deep-learning approach. \textit{Supervised learning} involves using a labeled dataset to train a classification model that predicts labels for out-of-sample, unlabeled data. \textit{Deep learning}, a subset of machine learning, employs neural network models with a "deep" architecture (many layers), allowing them to model complex nonlinear relationships and patterns in large and unstructured data, such as text and images. Training deep neural networks from scratch requires vast datasets and substantial computational power.

When such resources are unavailable, \textit{transfer learning} becomes a practical solution. In transfer learning, we utilize models pre-trained on large datasets and adapt them for our specific task. Instead of retraining the entire model, which can be resource-intensive, we often fine-tune  pre-trained models. Fine-tuning involves adjusting the last layers of the network to better fit our task while keeping the earlier layers fixed or making only minor adjustments. This approach facilitates training with a smaller dataset and less computing power by modifying only the relevant parts of the model.

\begin{figure}[!htbp]
\vspace{5mm}    

    \centering
        \caption{Convolutional Neural Network}
    \label{cnn_scheme}
    \includegraphics[width=1\linewidth]{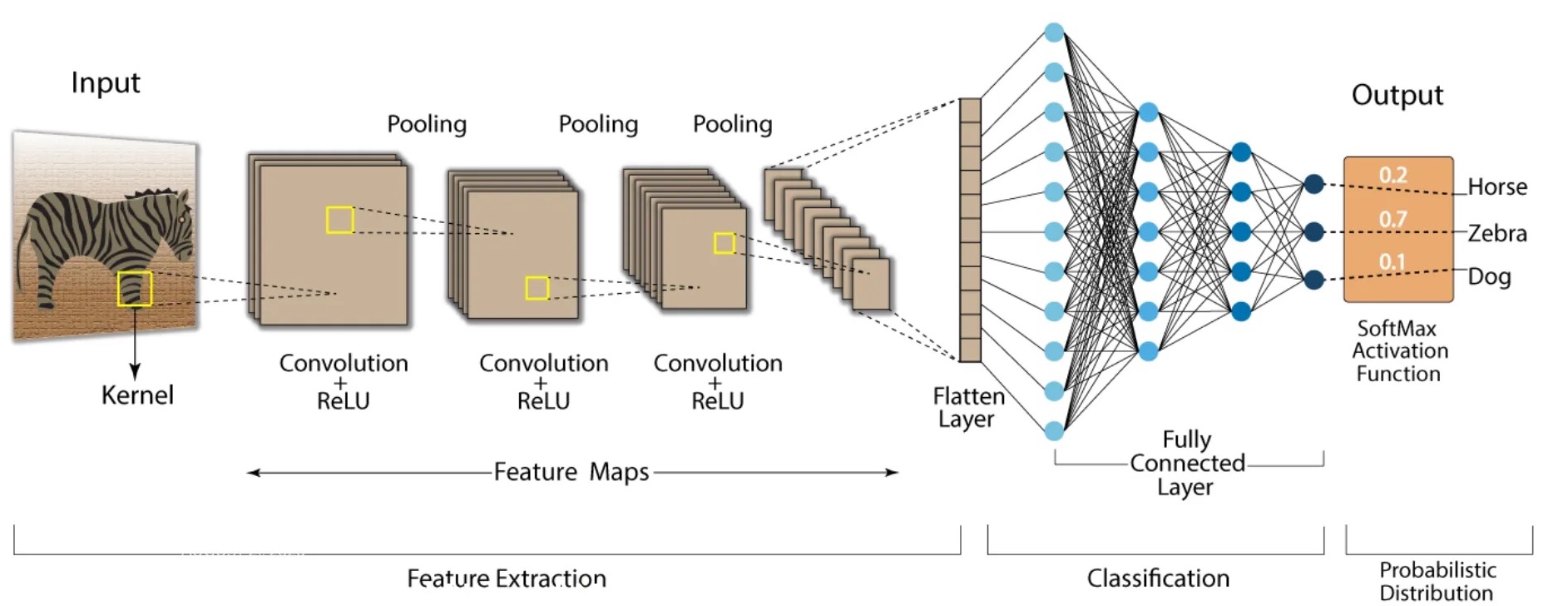}
    \begin{minipage}{15cm}
        \vspace{0.5cm}
        \footnotesize{\textit{Note:} Image is borrowed from a Medium article ``What is Convolutional Neural Network" by Nafiz Shahriar available \hyperlink{https://tinyurl.com/schCNN}{here}.}
    \end{minipage}
    \vspace{5mm}    

\end{figure}

In this study, we work with unstructured image data and employ a deep learning framework for image classification. Specifically, we use labels from our 816 images to train a classification model that automatically predicts visual sentiment labels for out-of-sample images. The state-of-the-art approach to image classification involves using convolutional neural networks\footnote{Original architecture of convolutional neural networks was introduced by \cite{lecun1989backpropagation}.}(CNNs). CNNs are composed of an input layer (which processes the image), multiple hidden layers, and an output layer (illustrated in Figure \ref{cnn_scheme}).

The primary distinction of CNNs from other neural networks lies in their use of convolutional layers in the hidden layers. In these layers, convolutional kernels (filters) slide across the input matrix to produce a feature map, which then serves as the input for subsequent layers of the network. The input to the network is a tensor (a multidimensional array) with the shape:
$$
\text{(number of inputs/images) x (input height) x (input width) x (number of input channels).} 
$$
The number of channels in an image depends on whether it is black-and-white or colored: a black-and-white image has only one channel, while a colored image, represented in the RGB spectrum, has three channels. We resize all images to a dimension of 224x224 pixels. For RGB images, the input data shape is (N, 224, 224, 3), where N is the number of images in the training set.

\begin{figure}[!htbp]
    \vspace{5mm}    
\centering
\caption{Feature Maps Visualization Heatmap (example with DenseNet-169)}
\label{img_transform_dn169}
\begin{subfigure}[b]{.52\textwidth}
    \centering
    \includegraphics[width=0.995\linewidth]{3.jpeg}  
\end{subfigure}
 \begin{subfigure}[b]{0.4\textwidth}
        \centering

       \tikz {\node (a) {\includegraphics[width=0.87\linewidth]{DN169V3_linear_layer_conv1_conv.png}};
        \draw[decorate,line width=0.3mm,decoration={calligraphic brace, amplitude=0.2cm}] (a.north east)+(0,-6pt) -- ([shift={(0, 6pt)}]a.south east)
        node[pos=0.5,right=5pt,black, font=\scriptsize]{Conv1};}
        \tikz {\node (b) {\includegraphics[width=0.87\linewidth]{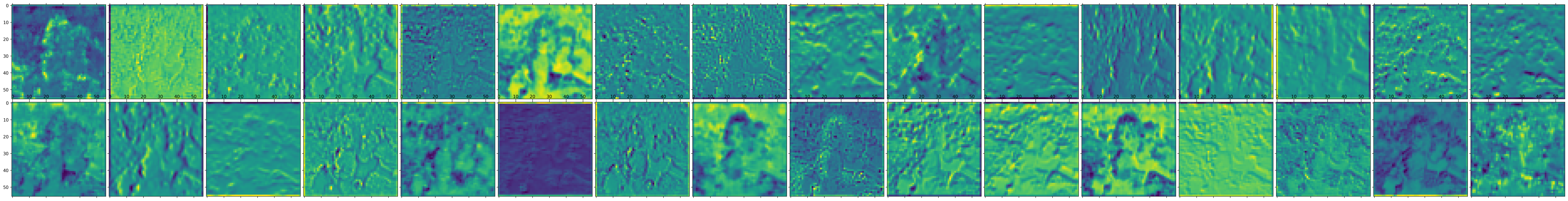}};
       \draw[decorate,line width=0.3mm,decoration={calligraphic brace, amplitude=0.1cm}] (b.north east)+(0,-6pt) -- ([shift={(0, 6pt)}]b.south east)
       node[pos=0.5,right=5pt,black, font=\scriptsize]{Dense1};}
        \tikz {\node (c) {\includegraphics[width=0.87\linewidth]{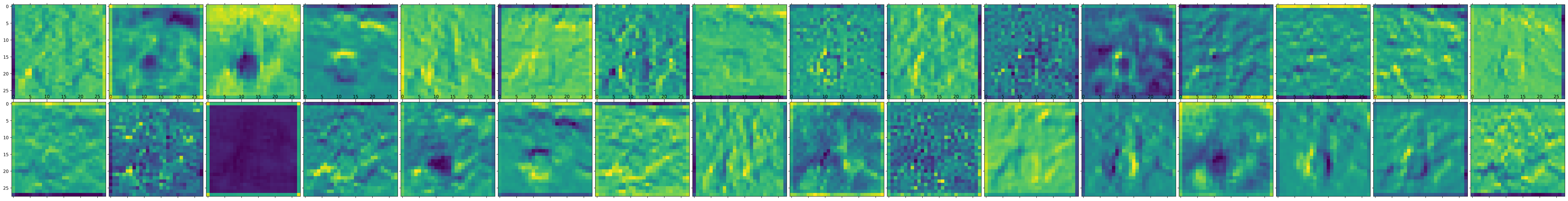}};
       \draw[decorate,line width=0.3mm,decoration={calligraphic brace, amplitude=0.1cm}] (c.north east)+(0,-6pt) -- ([shift={(0, 6pt)}]c.south east)
       node[pos=0.5,right=5pt,black, font=\scriptsize]{Dense2};} 
        \tikz {\node (d) {\includegraphics[width=0.87\linewidth]{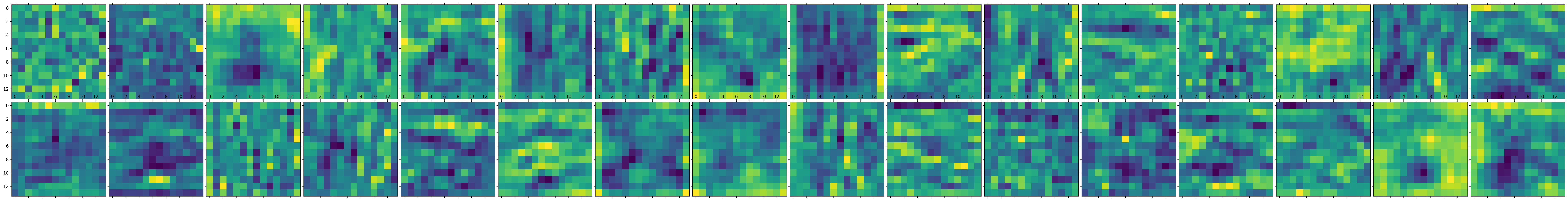} };
       \draw[decorate,line width=0.3mm,decoration={calligraphic brace, amplitude=0.1cm}] (d.north east)+(0,-6pt) -- ([shift={(0, 6pt)}]d.south east)
       node[pos=0.5,right=5pt,black, font=\scriptsize]{Dense3};} 
         \tikz {\node (e) {\includegraphics[width=0.87\linewidth]{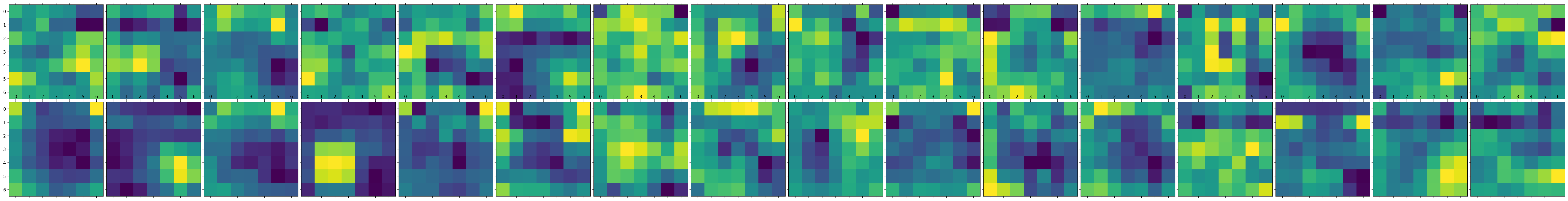} };
       \draw[decorate,line width=0.3mm,decoration={calligraphic brace, amplitude=0.1cm}] (e.north east)+(0,-6pt) -- ([shift={(0, 6pt)}]e.south east)
       node[pos=0.5,right=5pt,black, font=\scriptsize]{Dense4};}
    \end{subfigure}
    \begin{minipage}{15cm}
        \vspace{0.3cm}
        \footnotesize{\textit{Note:} The left side displays the original image that was input into the neural network, while the right side illustrates the image transformations within the network. Each block represents the resulting image transformation at the first convolutional layer and at four dense blocks of the DenseNet-169. Specifically, Conv1 corresponds to the second layer of the original network, and Dense1 through Dense4 correspond to layers 12, 58, 146, and 374 of the original network, respectively.}
    \end{minipage}
        \vspace{5mm}    
\end{figure}

After passing through the convolutional layers, an image is transformed into a feature map—a more abstract representation with the shape of:
$$
\text{(number of inputs/images) × (feature map height) × (feature map width) × (feature map channels).}
$$
Figure \ref{img_transform_dn169} illustrates feature maps at various levels of the CNN, showing how the image transforms as it progresses through the network’s depth. 

Besides convolutional layers CNNs also include pooling layers, which reduce the data dimensions by aggregating outputs from groups of neurons in one layer into single neurons in the next layer. Finally, CNNs conclude with fully connected (dense) layers, where each neuron in a layer connects to every neuron in the preceding layer. The network’s output is then passed through a softmax activation function in the final layer:

$$
\sigma(z)_i = \frac{e^{z_i}}{\sum^{K}_{j=1} e^{z_j}},
$$
(where $z_i$ is the i-th element of the output of the fully connected layer) to produce probabilities of being in each of the predicted classes. 

\subsubsection*{Fine-tuning}
In this study, given the relatively small size of our image dataset, we employ transfer learning by fine-tuning pre-trained deep convolutional neural networks. We tested three well-established convolutional neural networks commonly used for visual sentiment analysis: ResNet50V2 \citep{he2016identity}, DenseNet-121, and DenseNet-169 \citep{huang2017densely}, all pre-trained on the large ImageNet dataset. During fine-tuning, we adapted these models to our specific task by gradually 'unfreezing' and retraining several of the last layers or blocks of layers.

For each network, we implemented three fine-tuning strategies: 1) using the weights from the entire baseline model and only training the final fully connected layers responsible for classification; 2) retraining one of the last convolutional blocks with additional pooling layers (or a dense block with an additional transition layer for DenseNets) and adding newly trained fully connected layers for classification; 3) retraining two of the last convolutional blocks with additional pooling layers (or dense blocks with additional transition layers for DenseNets) and including newly trained fully connected layers for classification.

\begin{figure}[!htbp]
\vspace{5mm} 
    \centering
    \caption{Model Architectures}
    \label{architectures}
    \begin{subfigure}[b]{\textwidth}
        \centering
      \caption{Retraining Baseline DenseNet-121 with Additional Layers}
    \label{DN121}
   \resizebox{14cm}{!}{ \tikz {\node (a) {\includegraphics[width=\textwidth]{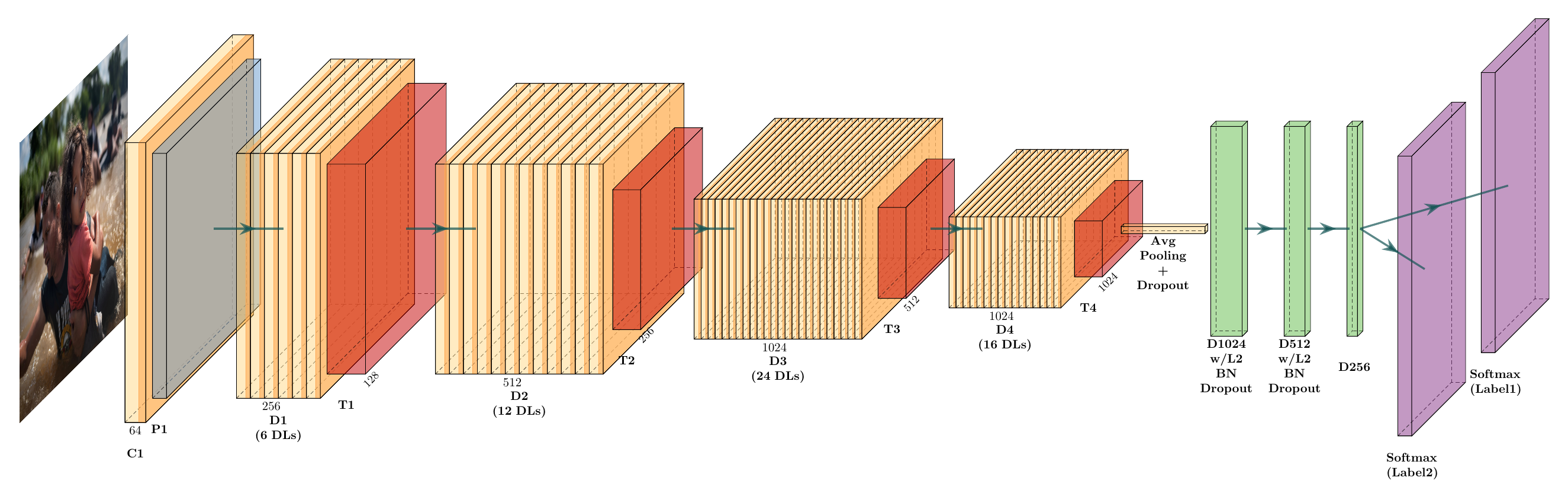}};
        \draw[decorate,line width=0.4mm,decoration={calligraphic brace, amplitude=0.2cm}] (a.south east)+(0, 0cm) -- ([shift={(12cm, 0cm)}]a.south west)
        node[pos=0.5, black, below=5pt, font=\tiny]{V1}; 

        \draw[decorate,line width=0.4mm,decoration={calligraphic brace, amplitude=0.2cm}] (a.south east)+(0, -0.5cm) -- ([shift={(10.2cm, -0.5cm)}]a.south west)
        node[pos=0.5, black, below=5pt, font=\tiny]{V2}; 
        
        \draw[decorate,line width=0.4mm,decoration={calligraphic brace, amplitude=0.2cm}] (a.south east)+(0, -1cm) -- ([shift={(7.5cm, -1cm)}]a.south west)
        node[pos=0.5, black, below=5pt, font=\tiny]{V3};
        
        \draw[decorate,line width=0.5mm,decoration={calligraphic brace, amplitude=0.2cm}] (a.north west)+(2cm, 0) -- ([shift={(-4.5cm, 0)}]a.north east)
        node[pos=0.5, black, above=5pt, font=\scriptsize]{Baseline Model};} 
        }
        \vspace{1cm}
\end{subfigure}
     \begin{subfigure}[b]{\textwidth}
         \centering
      \caption{Retraining Baseline DenseNet-169 with Additional Layers}
    \label{DN169}
    \resizebox{14cm}{!}{
   \tikz {\node (b) {\includegraphics[width=\textwidth]{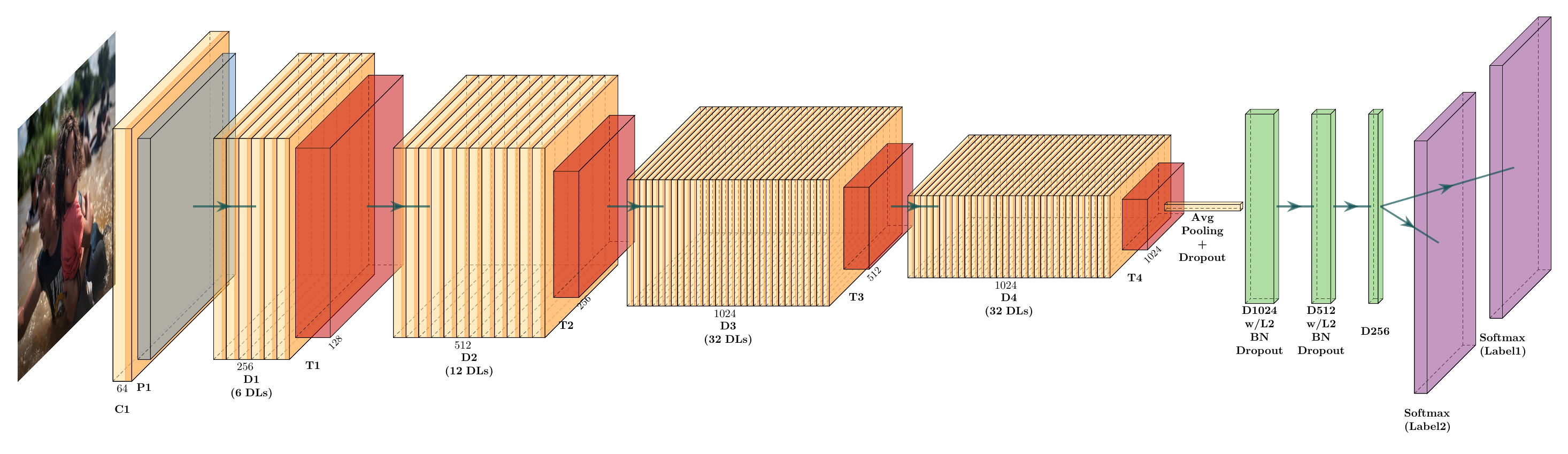}};
        \draw[decorate,line width=0.4mm,decoration={calligraphic brace, amplitude=0.2cm}] (b.south east)+(0, 0cm) -- ([shift={(12.4cm, 0cm)}]b.south west)
        node[pos=0.5, black, below=5pt, font=\tiny]{V1}; 

        \draw[decorate,line width=0.4mm,decoration={calligraphic brace, amplitude=0.2cm}] (b.south east)+(0, -0.5cm) -- ([shift={(9.6cm, -0.5cm)}]b.south west)
        node[pos=0.5, black, below=5pt, font=\tiny]{V2}; 
        
        \draw[decorate,line width=0.4mm,decoration={calligraphic brace, amplitude=0.2cm}] (b.south east)+(0, -1cm) -- ([shift={(6.7cm, -1cm)}]b.south west)
        node[pos=0.5, black, below=5pt, font=\tiny]{V3};
        
        \draw[decorate,line width=0.5mm,decoration={calligraphic brace, amplitude=0.3cm}] (b.north west)+(2cm, 0) -- ([shift={(-4cm, 0)}]b.north east)
        node[pos=0.5, black, above=6pt, font=\scriptsize]{Baseline Model};} 
}
\end{subfigure}
\begin{minipage}{15cm}
\vspace{0.5cm}
    \footnotesize{\textit{Note:} 
The images illustrate the neural network architectures of the adapted convolutional neural networks (CNNs). Each model includes a baseline configuration, which was initially downloaded using Keras libraries and pre-trained on the ImageNet dataset. The versions V1, V2, and V3 represent different re-training exercises for the CNN, showing the parts of the base model that have been updated and highlighting additional layers introduced between the base model and the final classification layer. While the schematic visualization depicts a softmax layer as the final classification layer, it is replaced with a linear activation function for the linear prediction task.}
\end{minipage}
\vspace{5mm} 
\end{figure}

In each fine-tuning scenario, we incorporated Batch Normalization layers\footnote{A batch normalization layer is a layer that normalizes the activations (output of a neuron after applying a specific activation function) of the previous layer for each mini-batch. This normalization is done by scaling and shifting the activations to have a mean of zero and a standard deviation of one The primary purpose of batch normalization is to stabilize and accelerate the training process by reducing the internal covariate shift, which helps in making the network less sensitive to the initialization of weights and learning rates.} to stabilize and accelerate training, and Dropout layers\footnote{A dropout layer works by randomly setting a fraction of the input units to zero at each update during the training. Each time a batch of data is passed through the network, different sets of neurons are "dropped out" (ignored) and do not contribute to the forward pass or backpropagation. This reduces the likelihood of the network becoming overly dependent on specific neurons, leading to better generalization to out-of-sample data.} to prevent overfitting. The detailed network architectures are illustrated in Figure \ref{architectures}.

Each network was adapted for two main tasks: 1) multi-class classification with three categories (negative, neutral, and positive), and 2) linear prediction with sentiment labels measured on an interval scale [1,7]. We fine-tuned the models using the Python-based Keras API\footnote{Keras is an open-source, high-level neural networks API written in Python.} with our dataset of 816 labeled images, which were divided into training and validation sets with an 80/20 split. The training was conducted using the Adam optimizer\footnote{Adaptive Moment Estimation (Adam) adjusts the weights of the neural network to minimize the loss function, which measures the discrepancy between the network's predictions and the actual data.} with a learning rate of 0.0001\footnote{The learning rate is a hyperparameter that controls the size of the steps taken to update the model's weights during training.} over 50 epochs\footnote{An epoch is one complete pass through the entire training dataset. Training over multiple epochs allows the model to refine its weights and improve performance.} with a batch size of 32\footnote{Batch size refers to the number of training samples used to update the model's weights in each epoch.} and included an early stopping mechanism with a patience parameter of 10 epochs\footnote{Early stopping prevents overfitting by halting training when the validation loss does not improve for a specified number of epochs. The patience parameter determines how many epochs to wait for an improvement before stopping the training.}.

\subsubsection*{Dual-Task Classification}

The main contribution of our approach lies in leveraging systematic attitudinal differences among respondents to generate multiple labels for each image, instead of relying on a single label. This method enables us to produce a set of labels that reflect the perspectives of different respondent groups, specifically highlighting partisan differences in our study. Consequently, we derive a pair of sentiment labels for each image based on separate evaluations from Democrats and Republicans, $L_{D}$ and $L_{R}$, respectively, where $L_{D}$ represents the label from Democrats' evaluations and $L_{R}$ from Republicans' evaluations.

Multi-task learning is central to our method, involving the training of a single model to perform multiple tasks simultaneously. During training, the model’s initial layers are shared across tasks, learning common features from the input image data. The later layers, however, are specialized for each task to enhance performance. Each task has its own loss function, and the overall loss is computed as a weighted sum of the individual losses for each task. This approach leverages shared representations to improve performance across tasks compared to training separate models for each task \citep{crawshaw2020multi}.

The advantage of this multi-task learning approach lies in its computational efficiency. By enabling the model to concurrently learn features relevant to both tasks, it eliminates the necessity of training two separate models, despite the network's increased architectural complexity.

\section*{Results}

Tables \ref{nn_res_pos}-\ref{nn_res_harm} summarize the results for all trained models. Since we identified a significant partisan gap in labeling "Sentiment" and "Subject of Harm," to address it we focused on the Democratic and Republican labeling.

For linear prediction models, we report two metrics: mean absolute error: $\text{MAE} = \frac{1}{n} \sum_{i=1}^{n} \left| y_i - \hat{y}_i \right|$ and mean squared error: $\text{MSE} = \frac{1}{n} \sum_{i=1}^{n} (y_i - \hat{y}_i)^2$, where  $y_i$ is the actual value for the $i$-th data point, and $\hat{y}_i $ is the predicted by the neural network value for the $i$-th data point.

For multi-class classification, we report a weighted F1 score and accuracy score. The weighted F1 score for multi-class classification is calculated by considering the F1 score of each class, weighting it by the number of true instances for that class (support), and then averaging these scores. To calculate F1 score of each class, we calculate 
\begin{itemize}
    \item $\text{Precision} = \frac{\text{True Positives}}{\text{True Positives} + \text{False Positives}}$
    \item $\text{Recall} = \frac{\text{True Positives}}{\text{True Positives} + \text{False Negatives}}$
\end{itemize}
Then we combine precision and recall to calculate F1 score: 
$$\text{F1 Score} = 2 \times \frac{\text{Precision} \times \text{Recall}}{\text{Precision} + \text{Recall}}$$
Finally, we calculate the weighted F1 score as follows:
$$\text{Weighted F1 Score} = \frac{\sum_{i=1}^{N} \text{Support}_i \times \text{F1 Score}_i}{\sum_{i=1}^{N} \text{Support}_i}$$ where: $N$ is the number of classes, $\text{Support}_i$ is the number of true instances for class $i$, and $\text{F1 Score}_i$ is the F1 score for class $i$.

Accuracy score is calculated as: 
$$\text{Accuracy} = \frac{\text{Number of Correct Predictions}}{\text{Total Number of Predictions}}$$

The results presented in Table \ref{nn_res_pos} show that for the "Sentiment" outcome, the DenseNet-169 model was the best performer. In the linear prediction task, DenseNet-169 with all layers frozen achieved a mean absolute error of 0.51 for Democrats and 0.62 for Republicans. For the multi-class classification task, DenseNet-169 with layers frozen up to Dense Block 3 achieved a weighted F1 score of 0.7 and 83\% validation set accuracy for Democratic labels, while it achieved an F1 score of 0.62 and 64\% validation set accuracy for Republican labels.

\begin{figure}[tbh!]
\vspace{5mm} 
    \centering
    \caption{Confusion Matrices: "Sentiment"}
    \label{cm_pos}
 \begin{subfigure}[b]{0.45\textwidth}
 \centering 
 \caption{Labels by Democrats}
    \label{cm_pos_dem}
      \includegraphics[width=0.95\linewidth]{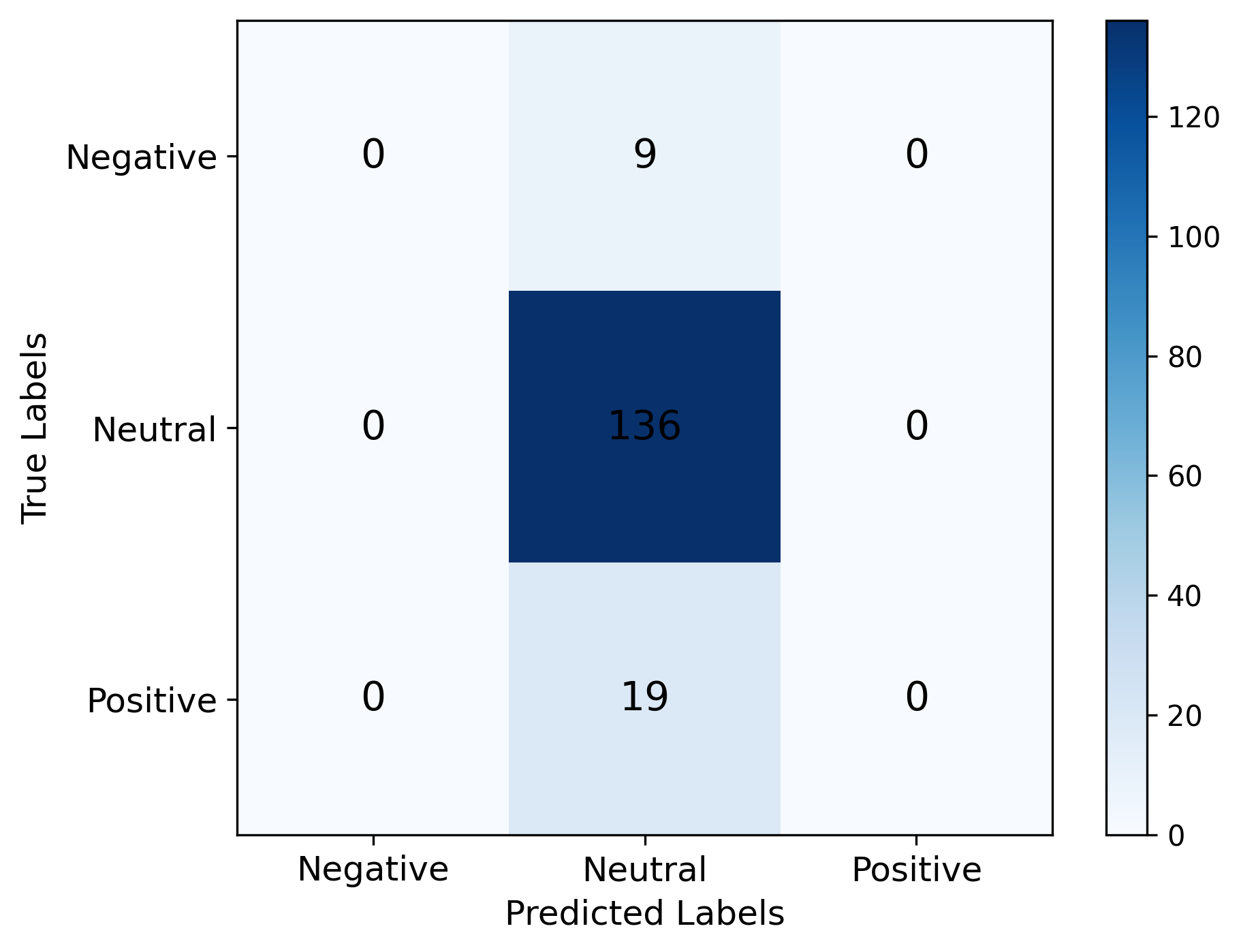}
  \end{subfigure}  
 \begin{subfigure}[b]{0.45\textwidth}
 \centering 
 \caption{Labels by Republicans}
    \label{cm_pos_rep}
      \includegraphics[width=0.95\linewidth]{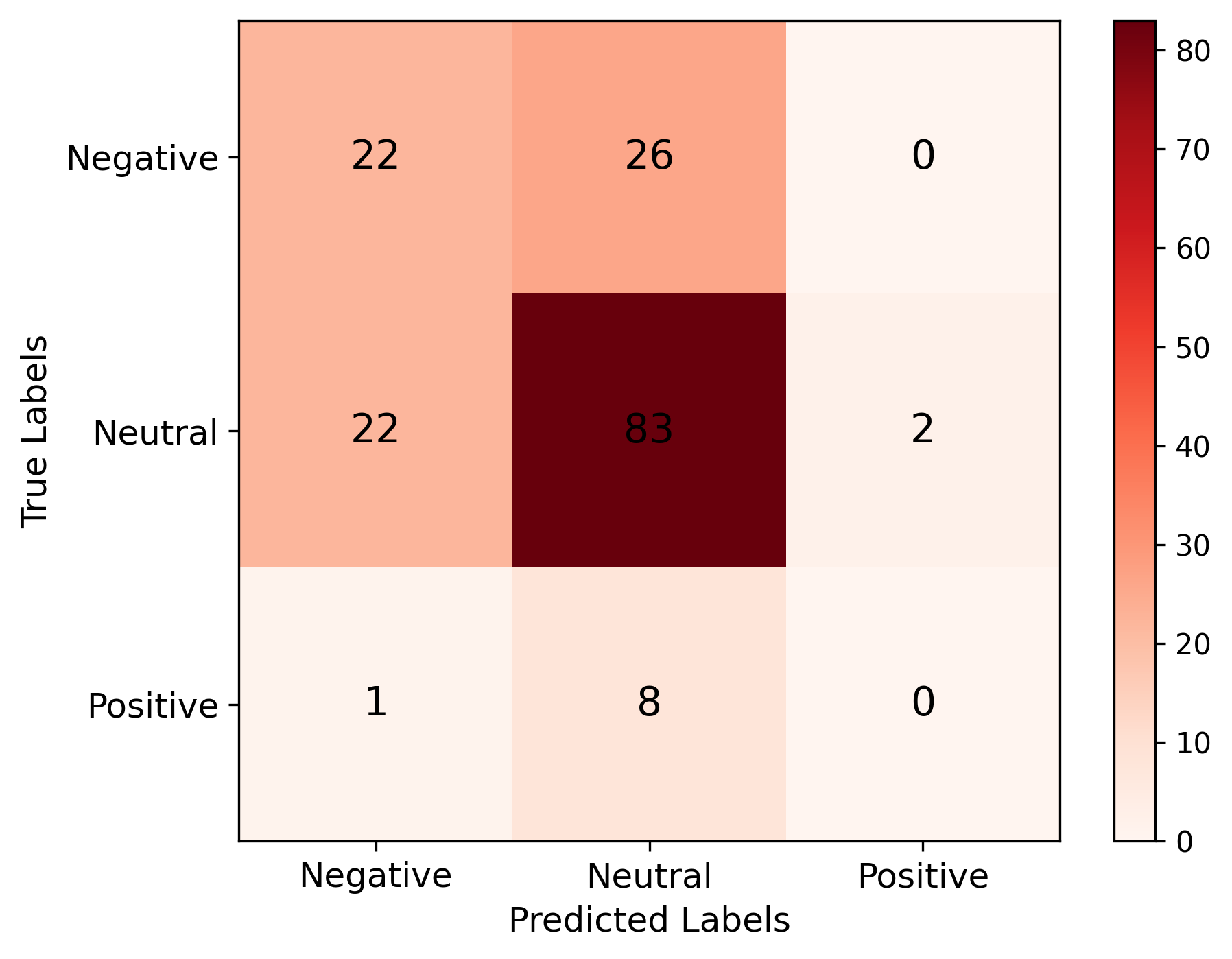}
  \end{subfigure}
\begin{minipage}{14cm}
\vspace{0.5cm}
    \footnotesize{\textit{Note:} Confusion matrices were constructed using the best-performing model for the "Sentiment" variable outcome, which was determined based on F-1 score and accuracy. This model, DenseNet-169, was enhanced by retraining the Dense 3 and Dense 4 blocks and incorporating additional dropout, batch normalization, and Dense layers. }
\end{minipage}
\vspace{5mm} 
\end{figure}

Figure \ref{cm_pos} shows the confusion matrix for the multi-class classification performed by the best model (DenseNet-169 in version three with retraining up to Dense Block 3). The model frequently confuses neutral with positive labels for Democratic labels, whereas it confuses neutral with negative labels for Republican labels.

Regarding the "Subject of Harm" outcome variable, as shown in Table \ref{nn_res_harm}, the DenseNet-121 model, with all layers frozen, demonstrated the best performance in the linear prediction task, achieving a mean absolute error of 0.54 for Democratic labels and 0.64 for Republican labels. In the multi-class classification task, despite similar performance across three models, the DenseNet-121 model, with layers frozen up to Dense Block 4, was selected. This model achieved a weighted F1 score of 0.7 and a validation set accuracy of 73\% for Democratic labels, and a weighted F1 score of 0.72 and a validation set accuracy of 76\% for Republican labels.

\begin{figure}[tbh!]
\vspace{5mm} 
    \centering
    \caption{Confusion Matrices: "Subject of Harm"}
    \label{cm_harm}
 \begin{subfigure}[b]{0.45\textwidth}
 \centering 
 \caption{Labels by Democrats}
    \label{cm_harm_rep}
      \includegraphics[width=0.95\linewidth]{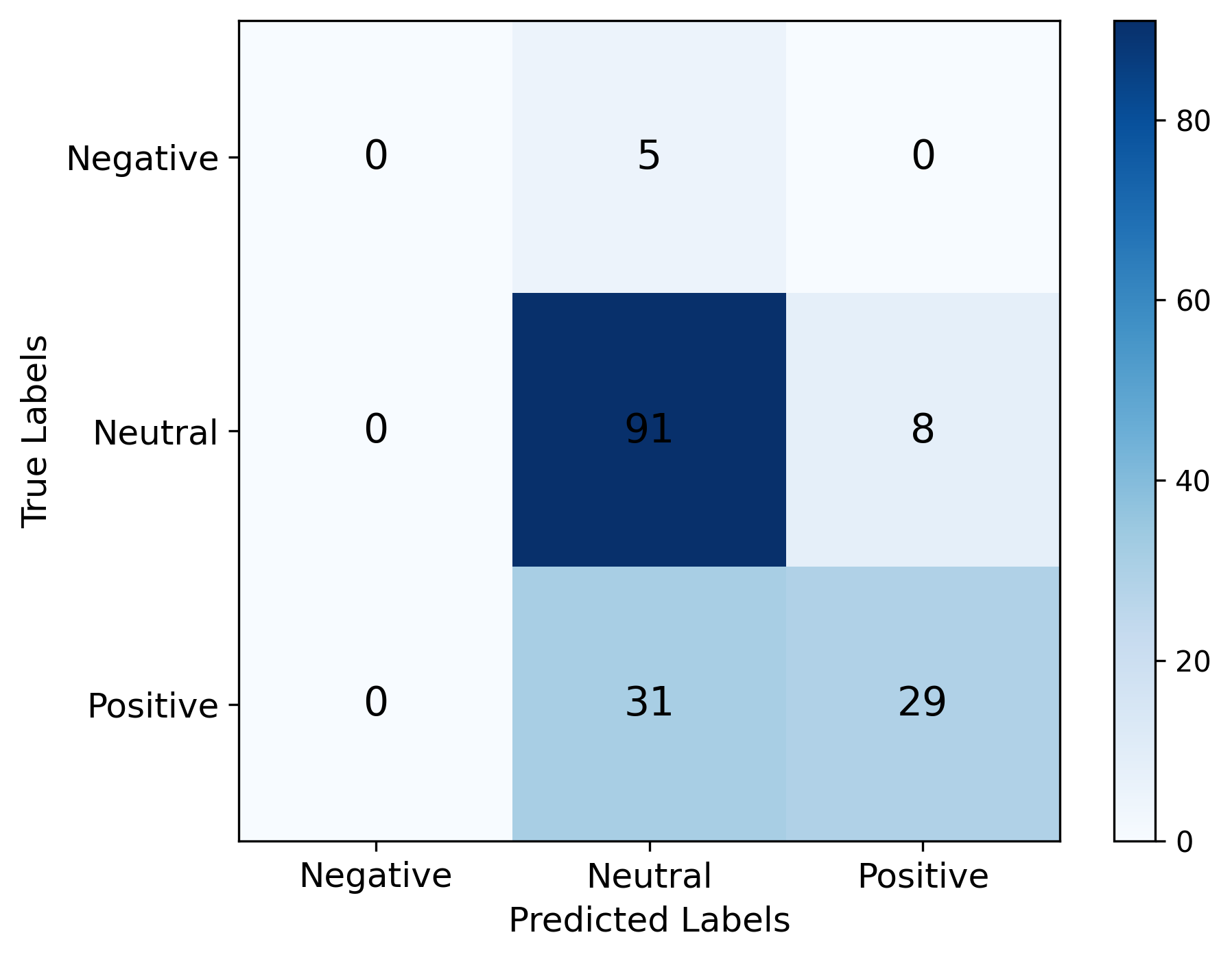}
  \end{subfigure}  
 \begin{subfigure}[b]{0.45\textwidth}
 \centering 
 \caption{Labels by Republicans}
    \label{cm_harm_rep}
      \includegraphics[width=0.95\linewidth]{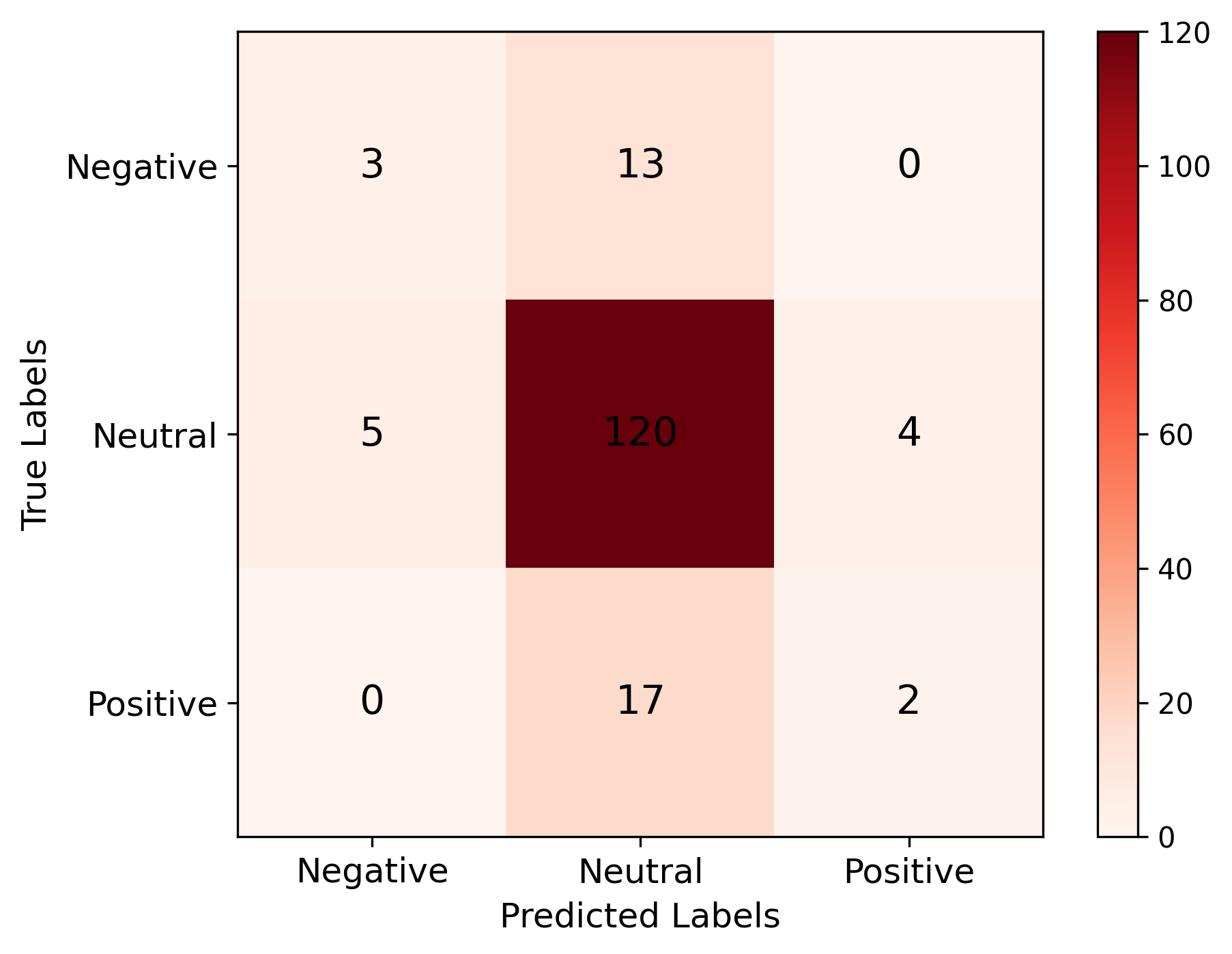}
  \end{subfigure}
\begin{minipage}{14cm}
\vspace{0.5cm}
    \footnotesize{\textit{Note:} Confusion matrices are built based on the best-performing model according to the F-1 score and accuracy for the "Objects of Harm" outcome - DenseNet-121 with retrained Dense 4 blocks and additional dropout, batch normalization, and Dense layers. }
\end{minipage}
\vspace{5mm} 
\end{figure}

Figure \ref{cm_harm} presents a confusion matrix for the multi-class classification task, constructed using the best-performing model (DenseNet-121 in version two of retraining). The confusion matrix shows that for labels produced by Democratic coders, the model primarily struggles to distinguish between neutral and positive labels. For Republicans, the confusion is also mainly between neutral and positive labels, with some additional confusion between negative and neutral labels.

\begin{table}[tbh!]
\vspace{5mm} 
\caption{Comparison of Neural Networks Performances for the "Sentiment" Outcome}
\label{nn_res_pos}
\begin{tabular}{ll|rr|rr|rr|rr|rr|rr}
\toprule
\hline
\multicolumn{2}{c|}{} & \multicolumn{6}{c|}{Linear Prediction} & \multicolumn{6}{c}{Multi-Class Classification}  \\
\hline
\multicolumn{2}{c}{} & \multicolumn{2}{|c}{RN50} & \multicolumn{2}{c}{DN121}& \multicolumn{2}{c}{DN169}& \multicolumn{2}{|c}{RN50} & \multicolumn{2}{c}{DN121}& \multicolumn{2}{c}{DN169}  \\
\hline
 V   & Label   &  MAE &  MSE &   MAE & MSE & MAE &   MSE &   F1 &   Acc &   F1 &   Acc &   F1 &   Acc \\
\midrule
 1   & Dem     &         0.58 &         0.59 &        0.5  &        0.41 &       \textbf{ 0.51 }&       \textbf{ 0.43 }&        0.75 &         0.79 &       0.76 &        0.82 &       0.76 &        0.83 \\ 
     & Rep     &         0.63 &         0.71 &        0.63 &        0.67 &       \textbf{ 0.62} &       \textbf{ 0.66} &        0.6  &         0.61 &       0.64 &        0.66 &       0.63 &        0.65 \\ \hline
 2   & Dem     &         0.66 &         0.75 &        0.52 &        0.48 &        1.19 &        1.77 &        0.77 &         0.82 &       0.78 &        0.83 &       0.76 &        0.83 \\
     & Rep     &         0.74 &         1    &        0.68 &        0.75 &        0.74 &        0.96 &        0.56 &         0.61 &       0.59 &        0.6  &       0.6  &        0.63 \\ \hline
 3   & Dem     &         0.64 &         0.64 &        0.59 &        0.56 &        0.62 &        0.64 &        0.75 &         0.83 &       0.76 &        0.83 &    \textbf{  0.75 }&      \textbf{ 0.83} \\
     & Rep     &         0.83 &         1.13 &        0.79 &        1.05 &        0.77 &        0.89 &        0.61 &         0.63 &       0.57 &        0.62 &      \textbf{ 0.62} &      \textbf{  0.64} \\
     \hline
     \hline
     \vspace{0.1cm}
\end{tabular}  
\begin{minipage}{16.5cm}
\footnotesize {\textit{Note:} 
Each column summarizes the results of individual neural network models: ResNet-50 V2, DenseNet-121, and DenseNet-169. The "V" denotes the version of re-training applied to these networks. Version 1 freezes all baseline layers, retraining only the classification or linear prediction layer. Version 2 retrains the last convolutional block (conv5) for ResNet-50 V2 and the last dense block (dense block 4) for DenseNet-121 and DenseNet-169, along with their corresponding classification and linear prediction layers. Version 3 retrains the last two convolutional blocks (conv4 and conv5) for ResNet-50 V2 and the last two dense blocks (dense blocks 3 and 4) for DenseNet-121 and DenseNet-169, along with their corresponding classification and linear prediction layers. For linear prediction models, we report Mean Absolute Error (MAE) and Mean Squared Error (MSE). For multi-class classification models, we report the F1 score with a weighted average across classes and accuracy scores for the validation set. The best-performing results are highlighted in bold.}
\end{minipage}
\vspace{5mm} 
\end{table}

\begin{table}[tbh!]
\vspace{5mm} 

\caption{Comparison of Neural Networks Performances for the "Subject of Harm" Outcome}
\label{nn_res_harm}
\begin{tabular}{ll|rr|rr|rr|rr|rr|rr}
\toprule
\hline
\multicolumn{2}{c|}{} & \multicolumn{6}{c|}{Linear Prediction} & \multicolumn{6}{c}{Multi-Class Classification}  \\
\hline
\multicolumn{2}{c}{} & \multicolumn{2}{|c}{RN50} & \multicolumn{2}{c}{DN121}& \multicolumn{2}{c}{DN169}& \multicolumn{2}{|c}{RN50} & \multicolumn{2}{c}{DN121}& \multicolumn{2}{c}{DN169}  \\
\hline
 V   & Label   &  MAE &  MSE &   MAE & MSE & MAE &   MSE &   F1 &   Acc &   F1 &   Acc &   F1 &   Acc \\
\midrule
  1   & Dem     &         0.64 &         0.63 &       \textbf{ 0.54} &        \textbf{0.5 } &        0.56 &        0.53 &        0.66 &         0.66 &       0.7  &        0.71 &       0.65 &        0.65 \\ 
     & Rep     &         0.6  &         0.58 &      \textbf{  0.62} &        \textbf{0.6  }&        0.63 &        0.63 &        0.69 &         0.7  &       0.71 &        0.72 &       0.71 &        0.76 \\ \hline
 2   & Dem     &         0.79 &         1.03 &        0.57 &        0.52 &        0.7  &        0.75 &        0.64 &         0.65 &      \textbf{ 0.7 } &     \textbf{   0.73} &      \textbf{ 0.72} &      \textbf{  0.73} \\
     & Rep     &         0.84 &         1.15 &        0.63 &        0.63 &        0.64 &        0.66 &        0.72 &         0.76 &      \textbf{ 0.72 }&        \textbf{0.76} &     \textbf{  0.71 }&       \textbf{ 0.73} \\  \hline
 3   & Dem     &         0.71 &         0.8  &        1.37 &        2.36 &        0.63 &        0.75 &        0.6  &         0.6  &       0.65 &        0.67 &     \textbf{  0.7}  &      \textbf{  0.72} \\
     & Rep     &         0.7  &         0.78 &        1.04 &        1.54 &        0.85 &        1.08 &        0.7  &         0.73 &       0.7  &        0.78 &      \textbf{ 0.71} &       \textbf{ 0.73} \\
     \hline
     \hline
     \vspace{0.1cm}
\end{tabular}  
\begin{minipage}{16.5cm}
\footnotesize {\textit{Note:} Each column summarizes the results of individual neural network models: ResNet-50 V2, DenseNet-121, and DenseNet-169. "V" represents the version of re-training applied to these models. Version 1 freezes all baseline layers, re-training only the classification or linear prediction layer. In Version 2, we re-trained the last convolutional block (conv5) for ResNet-50 V2 and the last dense block (dense block 4) for DenseNet-121 and DenseNet-169, including the corresponding classification and linear prediction layers. Version 3 involved re-training the last two convolutional blocks (conv4 and conv5) for ResNet-50 V2 and the last two dense blocks (dense blocks 3 and 4) for DenseNet-121 and DenseNet-169, along with their classification and linear prediction layers. For linear prediction models, we report the mean absolute error (MAE) and mean squared error (MSE). For multi-class classification models, we report the F1-score (weighted average across classes) and accuracy scores for the validation set. The best-performing results are indicated in bold.}
\end{minipage}
\vspace{5mm} 
\end{table}

\subsubsection*{Cross-Validation}
To validate the quality of the trained models, we performed K-fold cross-validation\footnote{K-fold cross-validation aims to ensure that the model's quality is not dependent on how the original set of images is split into training and validation sets. This procedure involves splitting the original dataset into $K$ batches and conducting $K$ different training tasks, each time using one of these subsets for validation against the remaining folds.} with 
$K=5$ for the best-performing models in linear and multi-class classification predictions for "Sentiment" and "Subject of Harm" outcomes. Since in the baseline training exercise we use 80/20 split for train/validation, here we also randomly split the dataset on 5 folds, where in each of the five iteration of cross-validation 1 of the 5 data folds (20\% of data) reserved for validation and the rest of the 4 folds (80\% of the data) is used for training.  Table \ref{kfcv} presents the results for K-fold cross-validation, averaging model performance metrics across all folds. Although the average metrics are slightly worse than the baseline results shown in Tables \ref{nn_res_pos}-\ref{nn_res_harm}, the accuracy levels for both "Sentiment" and "Subject of Harm" outcomes are similar to the baseline results. Additionally, we observe that the model performs better on "Sentiment" labels provided by Democrats, whereas for "Subject of Harm", the performance is better on labels provided by Republicans.

\begin{table}[tbh!]
\vspace{5mm} 
    \centering
        \caption{K-fold Cross Validation Results}
    \label{kfcv}
    \begin{tabular}{ll|rr|rr|rr|rr}
\toprule
\hline
\multicolumn{2}{c|}{} & \multicolumn{4}{c|}{Sentiment} & \multicolumn{4}{c}{``Subject of Harm''}  \\
\hline\multicolumn{2}{c|}{} & \multicolumn{2}{c|}{Linear} & \multicolumn{2}{c}{Multi-Class} & \multicolumn{2}{c|}{Linear} & \multicolumn{2}{c}{Multi-Class}  \\
\hline
 Fold   & Label   &      MAE &      MSE &       F1 &      Acc &   MAE &      MSE &       F1 &      Acc \\
\midrule
 1      & Dem     &    0.48 &    0.39 &   0.79 &    0.84 &    0.54 &    0.5  &   0.69 &    0.71 \\
     & Rep     &    0.61 &    0.6  &   0.6  &    0.61 &    0.6  &    0.55 &   0.68 &    0.68 \\ \hline
 2      & Dem     &    2.09 &    4.78 &   0.8  &    0.86 &    3.13 &   10.27 &   0.65 &    0.66 \\
     & Rep     &    1.75 &    3.78 &   0.59 &    0.64 &    2.69 &    7.79 &   0.68 &    0.72 \\ \hline
 3      & Dem     &    2.48 &    6.58 &   0.73 &    0.79 &    2.18 &    5.3  &   0.6  &    0.61 \\
     & Rep     &    1.68 &    3.4  &   0.64 &    0.69 &    2.06 &    4.79 &   0.72 &    0.74 \\  \hline
 4      & Dem     &    1.98 &    4.34 &   0.81 &    0.84 &    2.26 &    5.78 &   0.64 &    0.65 \\
     & Rep     &    1.7  &    3.57 &   0.64 &    0.69 &    2.31 &    6.04 &   0.69 &    0.79 \\ \hline
 5      & Dem     &    1.84 &    3.85 &   0.86 &    0.9  &    1.78 &    3.65 &   0.53 &    0.55 \\
     & Rep     &    1.66 &    3.35 &   0.54 &    0.63 &    1.57 &    3.2  &   0.63 &    0.72 \\\hline
     \hline
 \textbf{Avg}    & \textbf{Dem}     &    \textbf{1.77 }&    \textbf{3.99 }&  \textbf{ 0.8}  &    \textbf{0.85} &    \textbf{1.98 }&    \textbf{5.1  }&  \textbf{ 0.62 }&   \textbf{ 0.64} \\
   \textbf{Score}  & \textbf{Rep}     &   \textbf{ 1.48} &  \textbf{  2.94} &  \textbf{ 0.6  }&   \textbf{ 0.65} &   \textbf{ 1.85} &   \textbf{ 4.47 }&   \textbf{0.68} &   \textbf{ 0.73} \\\hline
 \hline
\end{tabular}
\begin{minipage}{12.5cm}
\vspace{0.3cm}
    \footnotesize{\textit{Note:} We performed 5-fold cross-validation using the best-performing model for each instance. Specifically, for the "Sentiment" and linear outcomes, we used DenseNet-169 V1; for the "Subject of Harm" and linear outcomes, we used DenseNet-121 V1; for multi-class classification with the "Sentiment" outcome, we used DenseNet-169 V3; and for multi-class classification with the "Subject of Harm" outcome, we used DenseNet-121 V2. For each fold, we report the mean absolute error (MAE) and mean square error (MSE) for linear prediction models, and the F1-score with weighted average across classes, along with accuracy scores for the validation set, for multi-class classification models. The average scores are calculated as simple averages of the reported metrics across the 5 folds.}
\end{minipage}
\vspace{5mm} 
\end{table}

\subsection*{Discussion}

Our study introduces and validates a novel approach for predicting visual sentiment in political images, where coders often have divergent attitudes. Here we address the limitations of traditional approaches and demonstrate the advantages of integrating our strategy into current sentiment analysis practices. Unlike conventional strategies that rely on large-scale labeling to mitigate low inter-coder agreement, our method incorporates the causes of this disagreement directly into the model training process. We argue that what is typically considered a limitation in traditional approaches actually provides valuable insights for identifying image sentiment.

Traditional visual sentiment analysis typically treats sentiment as a fixed attribute inherent to an image and determined solely by its visual content. Our study challenges this static view by presenting both theoretical and empirical evidence that visual sentiment should be understood as a reflection of individual perceptions and attitudes rather than an immutable property. This perspective highlights how sentiment is shaped by viewers' interpretations and responses, offering a more multifaceted view of how people perceive images.

The primary take-home message of our research is the significant and systematic impact of coders' social characteristics, such as political affiliations, on the labeling of visual sentiment. This factor should be considered when training visual sentiment models. Traditional methods often average responses to produce a single sentiment label, mistakenly assuming a uniform interpretation of visual content. 
%In contrast, our approach acknowledges the substantial and impactful variation in coders' preferences and perceptions, generating multiple labels that capture these diverse perspectives.

\begin{figure}[!htbp]
\vspace{5mm}
    \centering
    \caption{Predicted Sentiment Labels}
    \label{examples_sentiment}
    \begin{minipage}{0.46\textwidth}
        \centering
         \caption*{\textbf{Image A}}
        \label{}
        \includegraphics[width=0.9\textwidth]{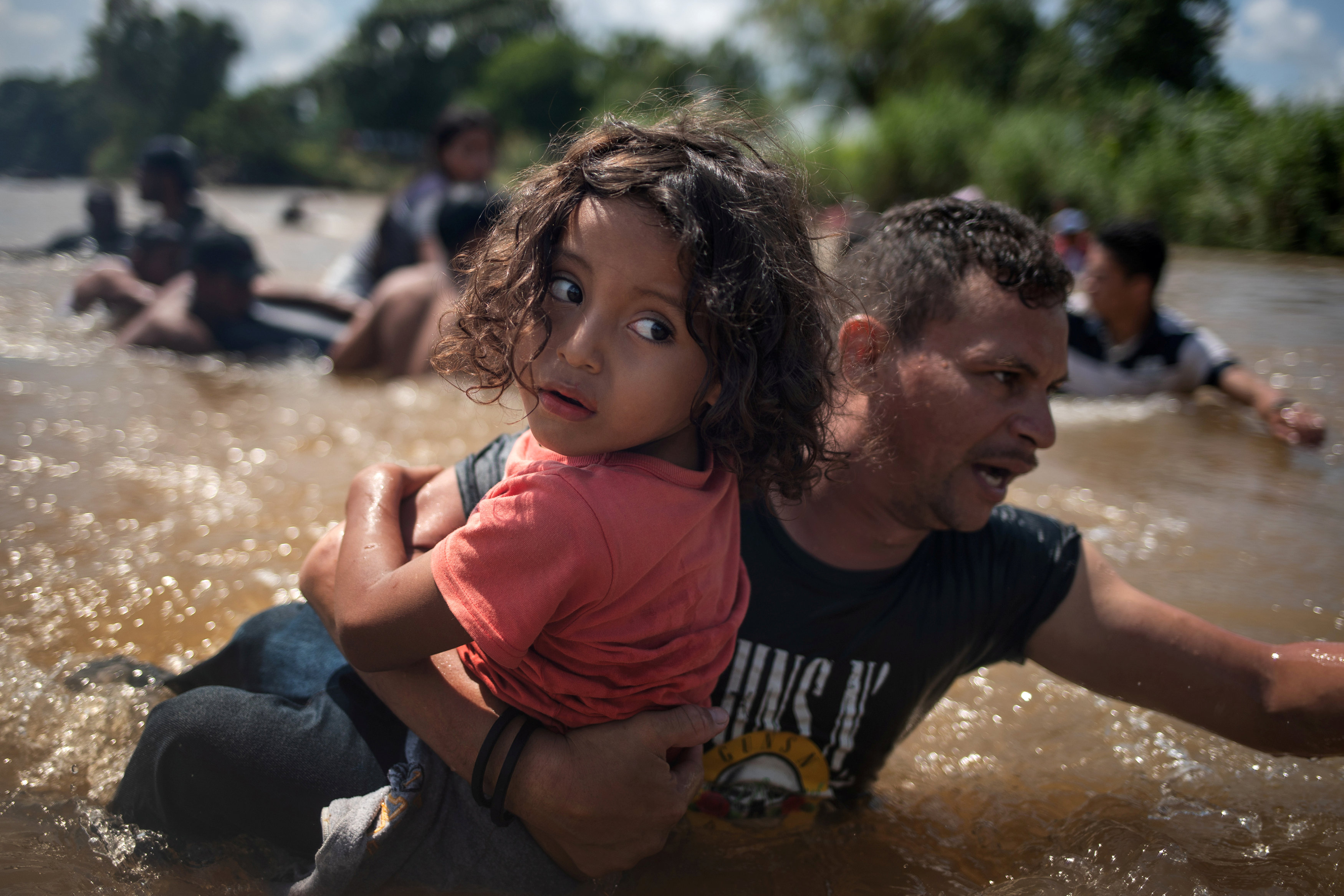}
 \footnotesize \textbf{``Sentiment''} Dem: [Neutral], Rep: [Negative]\\
        \footnotesize \textbf{``Subject of Harm''} Dem: [Positive], Rep: [Neutral]\\
        \vspace{0.2cm}
        \footnotesize \textbf{Conventional Labels}:\\
           "Sentiment": [Neutral], "Subject of Harm": [Neutral]\\
    \end{minipage}
    \hfill
    \begin{minipage}{0.46\textwidth}
        \centering
        \caption*{\textbf{Image B}}
        \label{}
        \includegraphics[width=\textwidth]{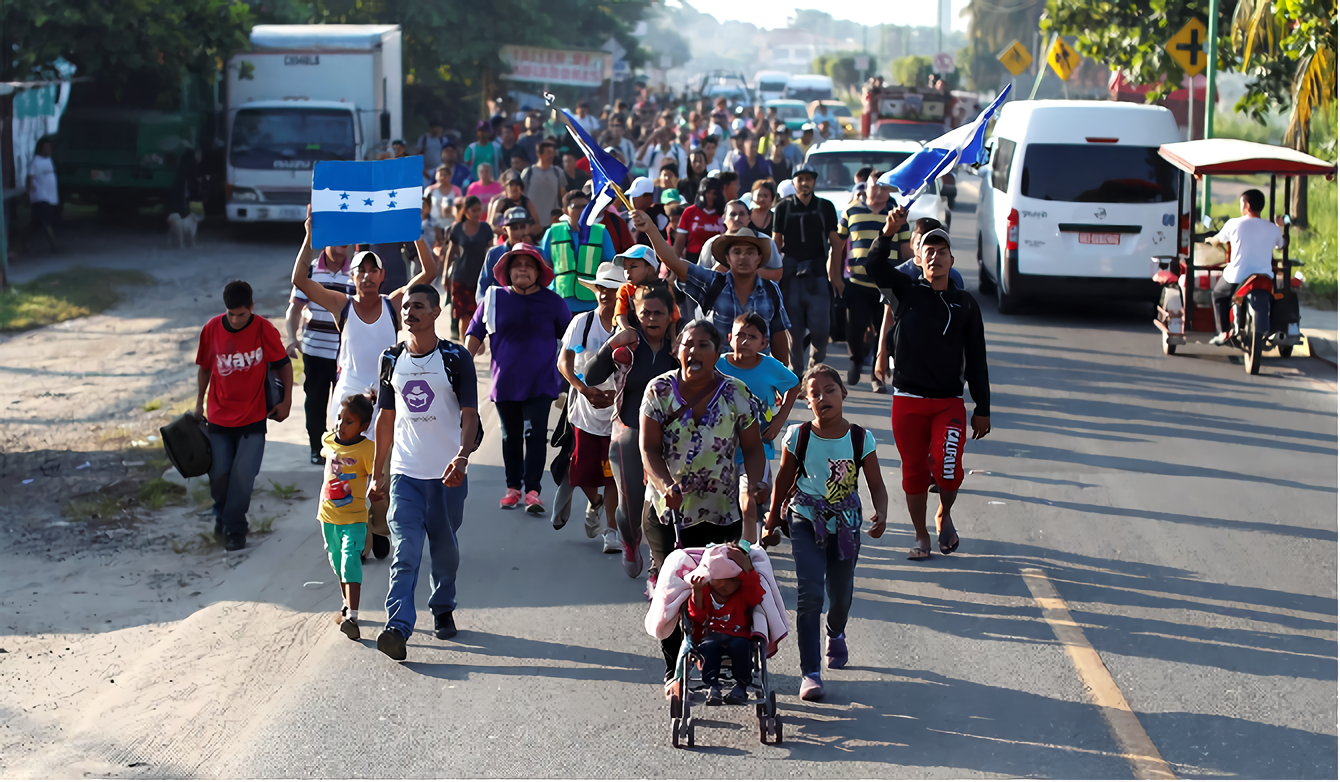}
        \footnotesize \textbf{``Sentiment''} Dem: [Neutral], Rep: [Negative]\\
        \footnotesize \textbf{``Subject of Harm''} Dem: [Positive], Rep: [Neutral]\\
        \vspace{0.2cm}
           \footnotesize \textbf{Conventional Labels}:\\
           "Sentiment": [Neutral], "Subject of Harm": [Positive]\\
    \end{minipage}
\vspace{5mm}   
\begin{minipage}{16cm}
\vspace{0.3cm}
    \footnotesize{\textit{Note:} "Sentiment" and "Subject of Harm" are used to proxy a broader concept of visual sentiment. If for "Sentiment" outcome negative, neutral, and positive labels are quite straightforward, for "Subject of Harm": negative label means that subjects were perceived as more dangerous and positive label means that subjects were perceived as more harmless. Conventional labels are based on average calculations across all the respondents without taking into account partisan divide in sentiment perceptions.}
\end{minipage}
\end{figure}

We advocate for a multi-label approach to visual sentiment analysis that captures the essential differences (in our case driven by partisan attitudes). Rather than consolidating diverse perspectives into a single label, our method assigns multiple labels to each image, reflecting the range of viewpoints influencing sentiment. Interpreting predictions simply as "our model predicted a positive/negative visual sentiment label for a given image" oversimplifies visual sentiment, which should reflect human attitudes in a more nuanced way. A more accurate interpretation is: "For group X (e.g., Democrats/Republicans), our model predicted a positive/negative visual sentiment label for this image." Although this approach may seem more complex, it provides a deeper understanding of how sensitive images are perceived by different social or ideological groups, leading to a more layered understanding of visual sentiment.

To illustrate this approach, we applied our trained models to the immigration-related images discussed in the introduction, each depicting different visual frames, as shown in Figure \ref{images12}. Using the conventional method, which averages all coders' evaluations, we would assign a "Neutral" label to both the "Sentiment" (average score of 4.1) and "Subject of Harm" (average score of 4.6) outcomes for image A and a "Neutral" label to the "Sentiment" (average score of 4.8) and 'Positive' label for "Subject of Harm" (average score of 5.6) outcomes for image B.
Our approach reveals a clear divergence in sentiment labels between Democrats and Republicans that would otherwise be overlooked. For the image of a man carrying a child, Democrats generally assign a more positive sentiment compared to Republicans. Similarly, for the image of a crowd marching, Democrats again show a more positive sentiment than Republicans, as illustrated in Figures \ref{examples_sentiment}.

Does this imply that we must evaluate every labeling task for the presence of cleavages, such as partisan divides, and account for them when building classification models and interpreting results? While it may not be necessary for every task, our approach demonstrates the importance of considering these divisions, especially with politically sensitive content. Researchers can utilize our model to identify whether an image's sentiment is subject to partisan splits, allowing for more accurate and context-aware analyses.

Furthermore, this study is particularly relevant for scholars studying political polarization in the US. For automatic label prediction, our trained model can forecast the sentiments that images elicit in different partisan groups. Additionally, for cases with polarizing sentiments\footnote{For instance, when Democrats assign a positive label (6 out of 7) while Republicans assign a negative label (2 out of 7).}, researchers can use the partisan sentiment labels to construct an index measuring the polarizing impact of an image: $|L_D - L_R|$, where $L_D$ and $L_R$ represent the sentiment labels from Democrats and Republicans, respectively. This index can provide valuable insights into the degree of polarization provoked by specific visual content.

\newpage

\bibliographystyle{apsr}
\bibliography{bib_annotator.bib}

\newpage
\appendix
\setcounter{table}{0}
\renewcommand\thetable{\Alph{section}.\arabic{table}}
\setcounter{figure}{0}
\renewcommand\thefigure{\Alph{section}.\arabic{figure}}
\section{Linear Prediction Quality}

\begin{figure}[!htbp]
    \centering
    \caption{Linear Prediction Quality: "Sentiment"}
    \label{lin_q_pos}
 \begin{subfigure}[b]{0.45\textwidth}
 \centering 
 \caption{Actual vs. Predicted "Sentiment" Evals. (Labels by Democrats)}
    \label{pred_pos_dem}
      \includegraphics[width=0.95\linewidth]{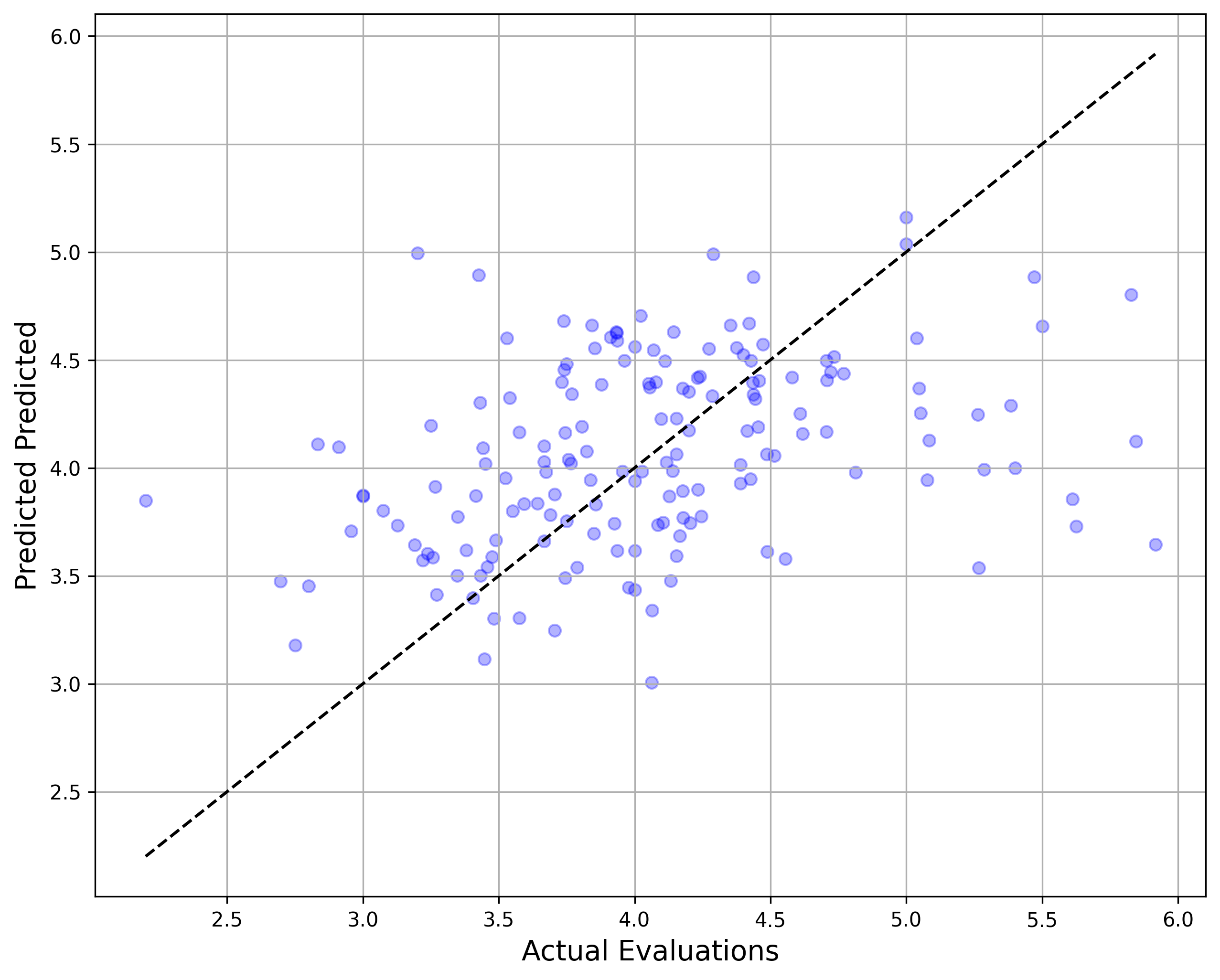}
            \vspace{0.3cm}
  \end{subfigure}  
  \hfill
 \begin{subfigure}[b]{0.45\textwidth}
 \centering 
  \caption{Actual vs. Predicted "Sentiment" Evals. (Labels by Republicans)}
    \label{pred_pos_rep}
      \includegraphics[width=0.95\linewidth]{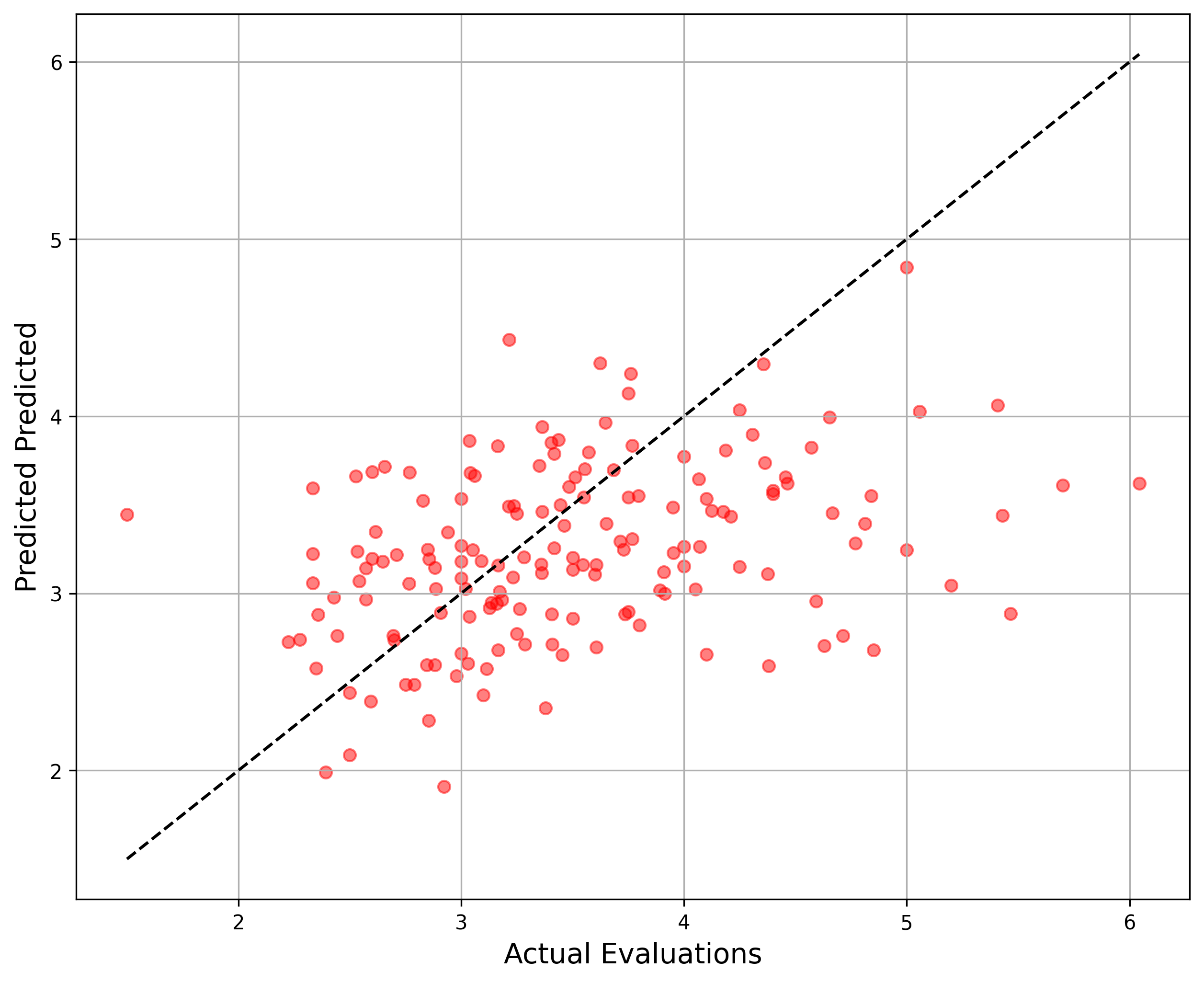}
      \vspace{0.3cm}
  \end{subfigure}
   \begin{subfigure}[b]{0.45\textwidth}
 \centering 
 \caption{Residuals of  "Sentiment" Predictions (Labels by Democrats)}
    \label{res_pos_dem}
      \includegraphics[width=0.95\linewidth]{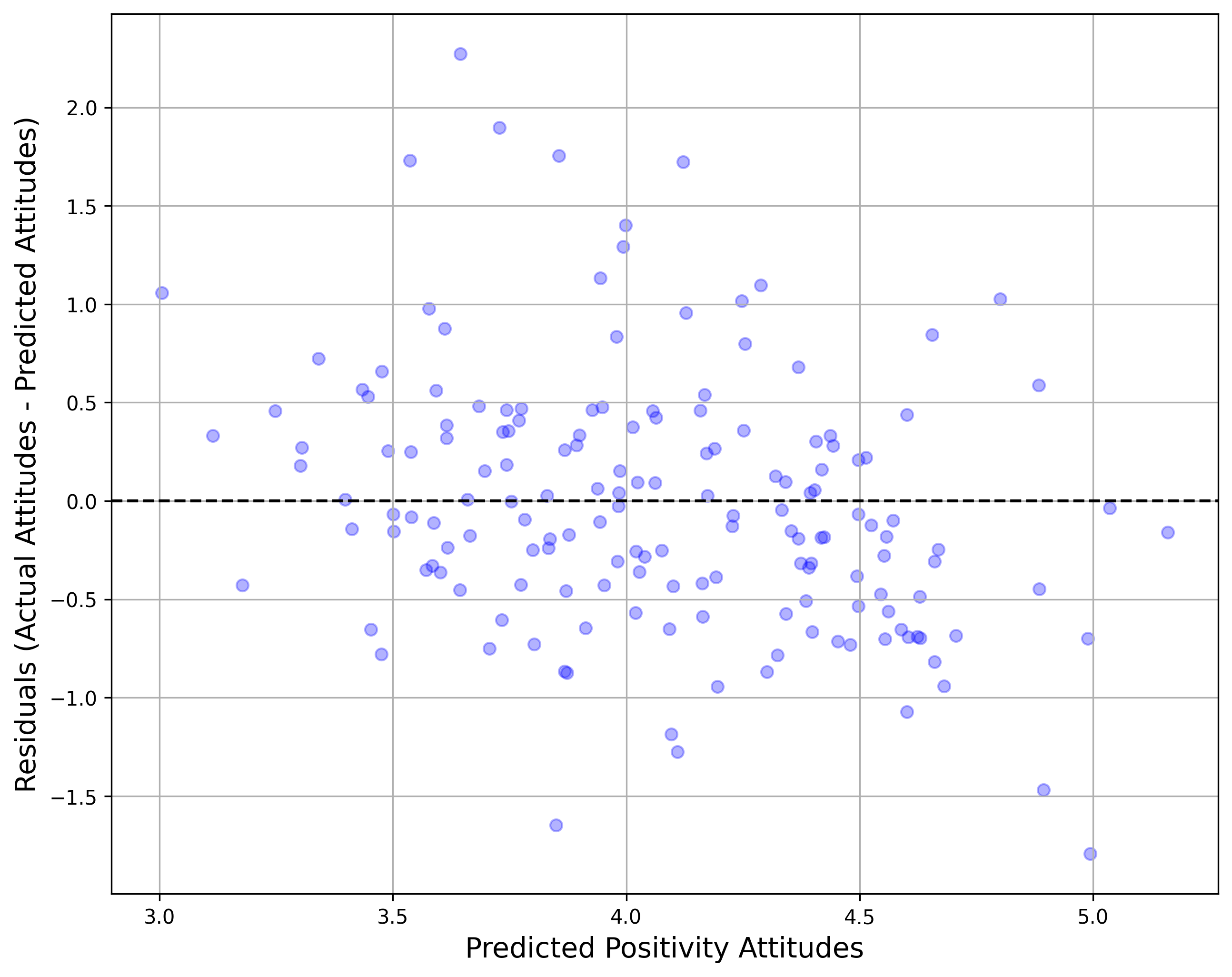}
  \end{subfigure}  
    \hfill
 \begin{subfigure}[b]{0.45\textwidth}
 \centering 
  \caption{Residuals of  "Sentiment" Predictions (Labels by Republicans)}
    \label{res_pos_rep}
      \includegraphics[width=0.95\linewidth]{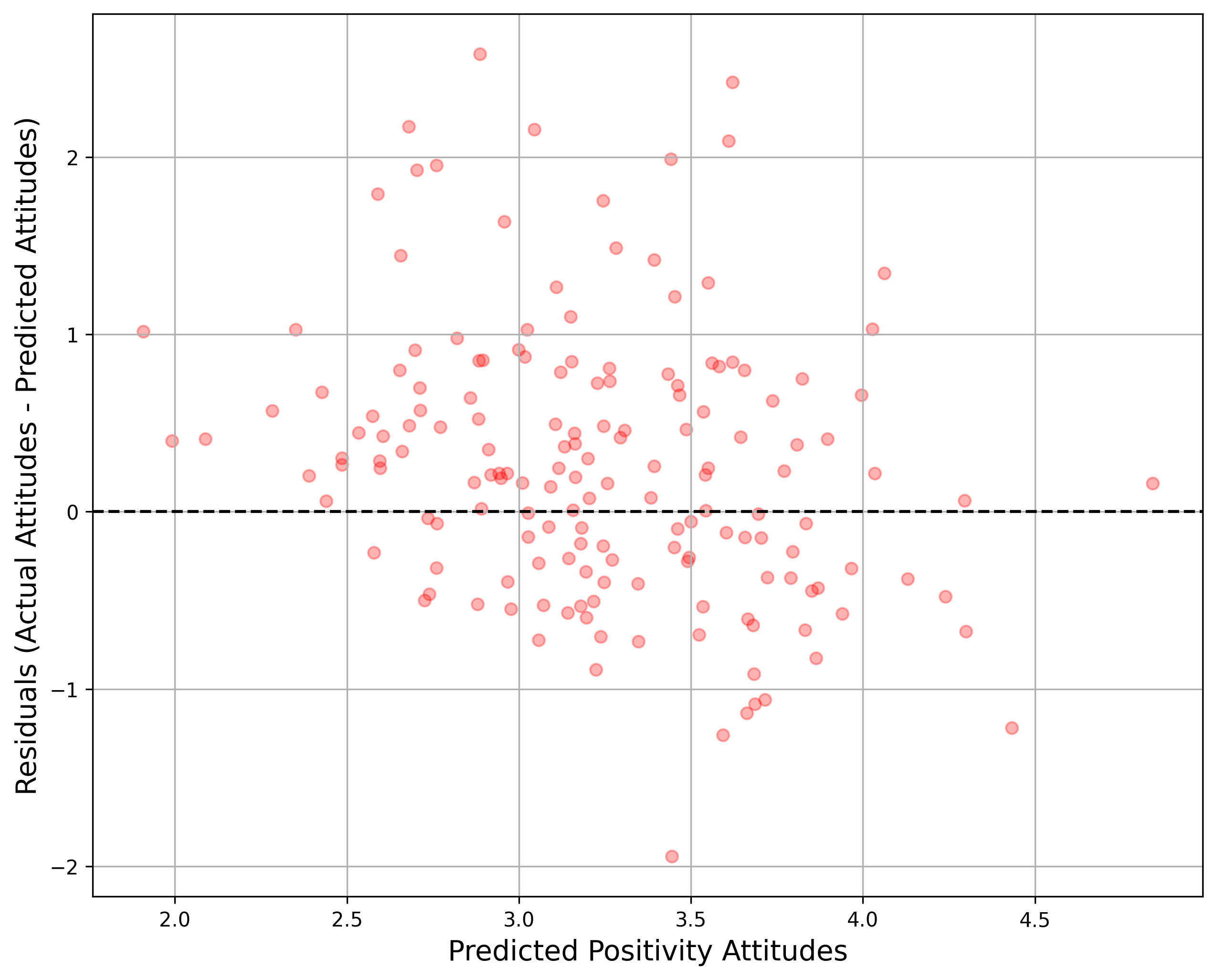}
  \end{subfigure}
\begin{minipage}{16cm}
\vspace{0.5cm}
    \footnotesize{\textit{Note:} The plots are generated using the prediction and residual values for the validation set for the "Sentiment" outcome estimated with the best-performing model: DenseNet-169 V1. This baseline model, without retraining, includes additional dropout, batch normalization, dense layers, and a linear prediction layer.}
\end{minipage}
\end{figure}

\begin{figure}[!htbp]
    \centering
    \caption{Linear Prediction Quality: "Subject of Harm"}
    \label{lin_q_harm}
 \begin{subfigure}[b]{0.45\textwidth}
 \centering 
 \caption{Actual vs. Predicted "Subject of Harm" Evals. (Labels by Democrats)}
    \label{pred_harm_dem}
      \includegraphics[width=0.95\linewidth]{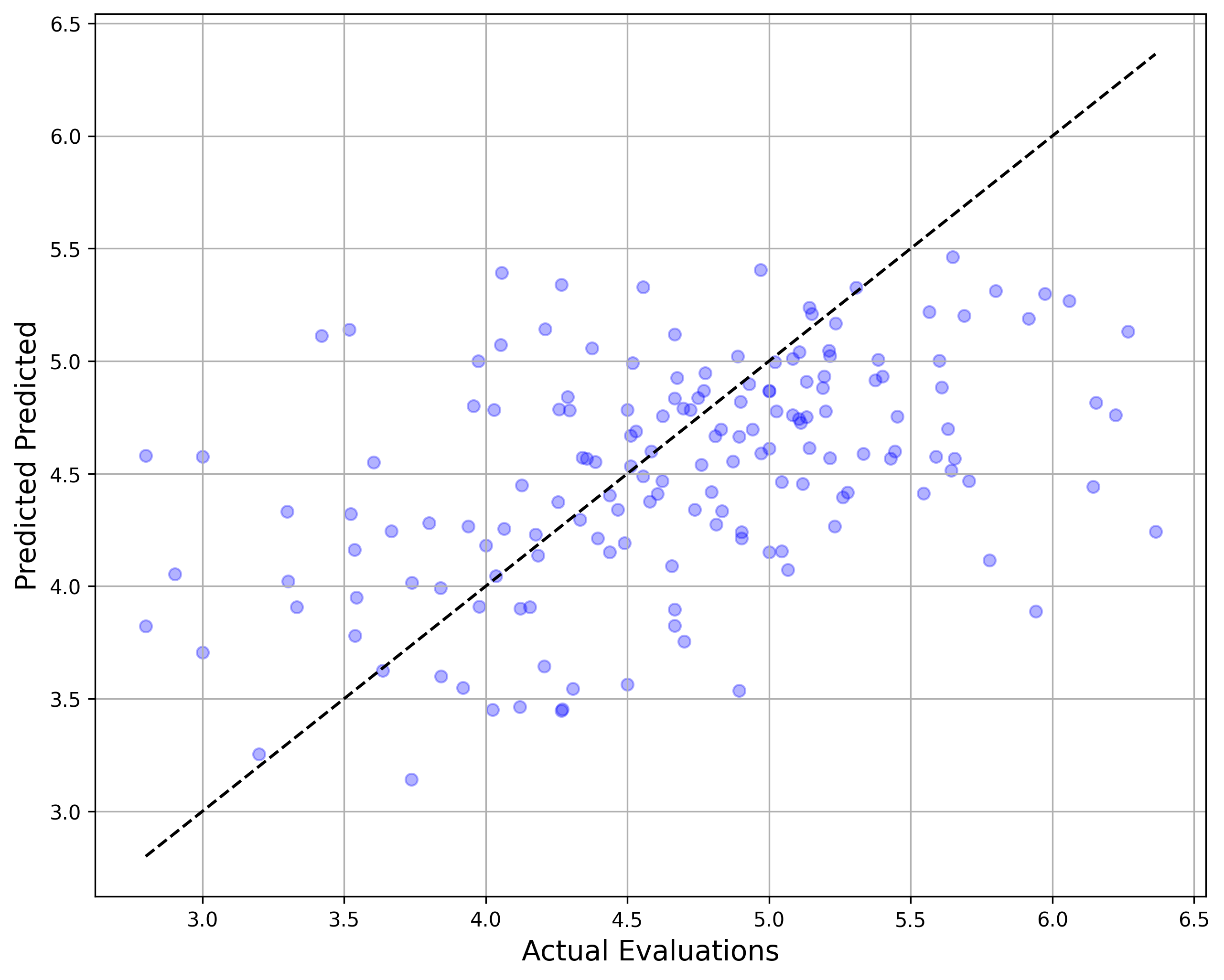}
            \vspace{0.3cm}
  \end{subfigure}  
  \hfill
 \begin{subfigure}[b]{0.45\textwidth}
 \centering 
  \caption{Actual vs. Predicted "Subject of Harm" Evals. (Labels by Republicans)}
    \label{pred_harm_rep}
      \includegraphics[width=0.95\linewidth]{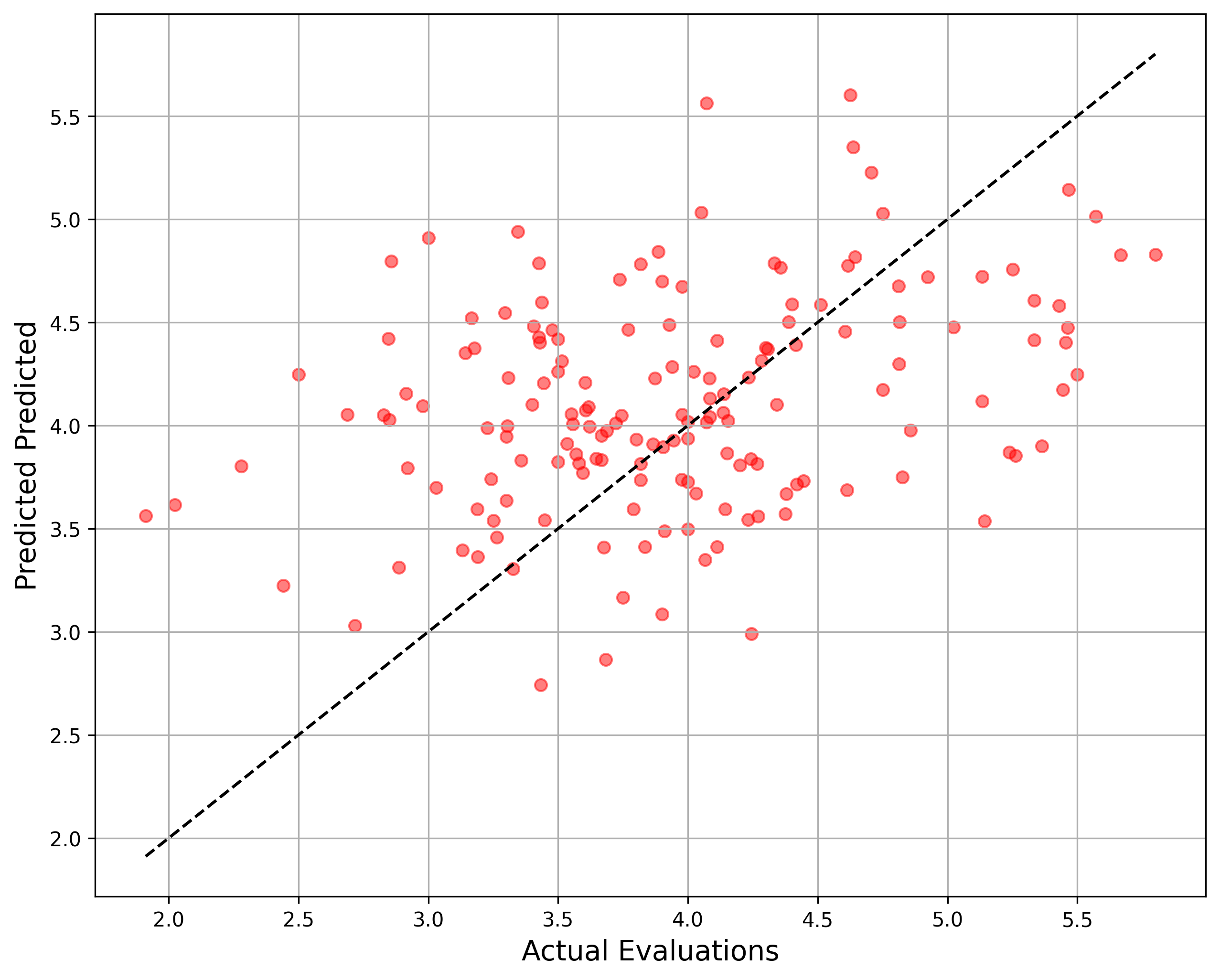}
      \vspace{0.3cm}
  \end{subfigure}
   \begin{subfigure}[b]{0.45\textwidth}
 \centering 
 \caption{Residuals of "Subject of Harm" Predictions  (Labels by Democrats)}
    \label{res_harm_dem}
      \includegraphics[width=0.95\linewidth]{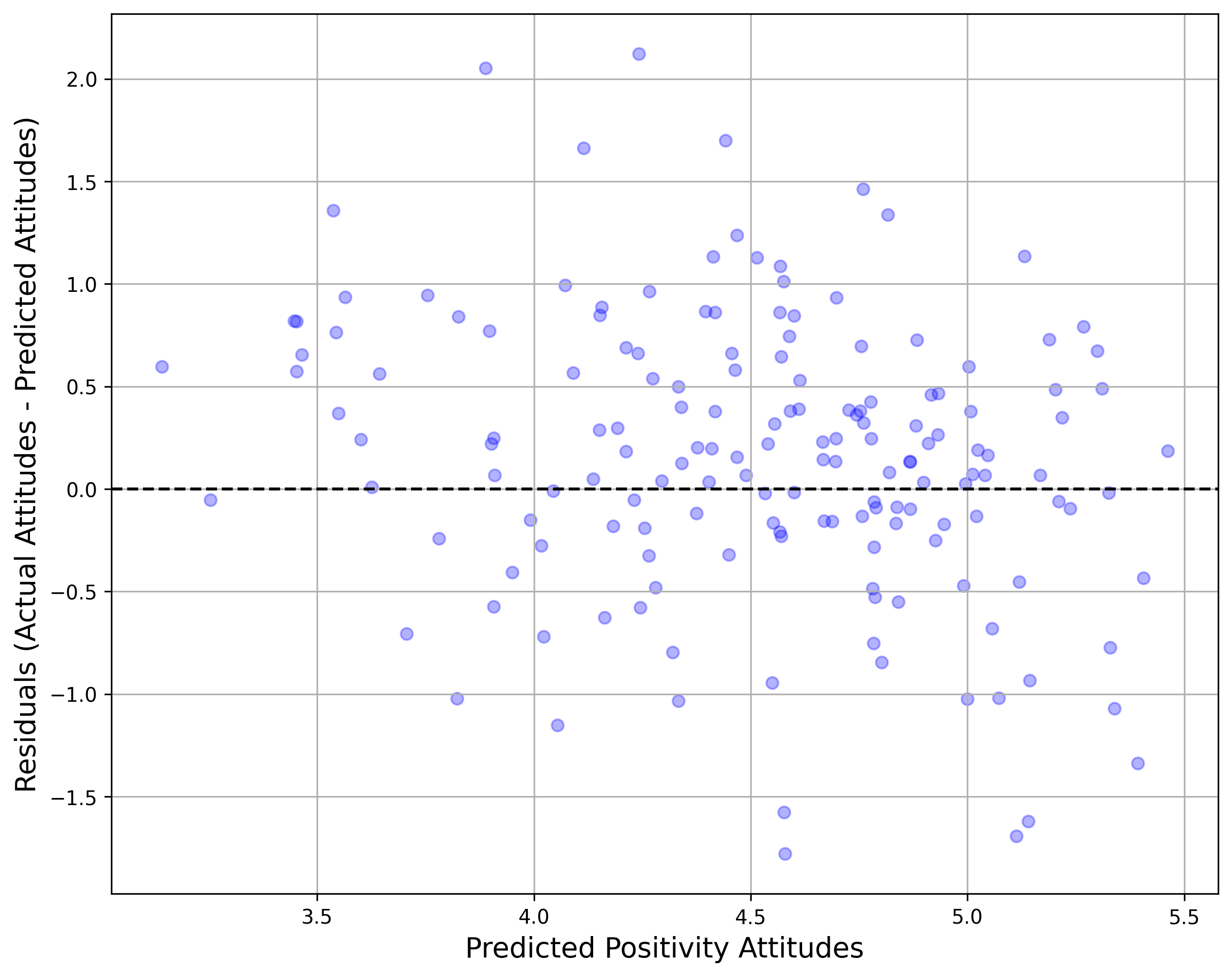}
  \end{subfigure}  
    \hfill
 \begin{subfigure}[b]{0.45\textwidth}
 \centering 
  \caption{Residuals of "Subject of Harm" Predictions (Labels by Republicans)}
    \label{res_harm_rep}
      \includegraphics[width=0.95\linewidth]{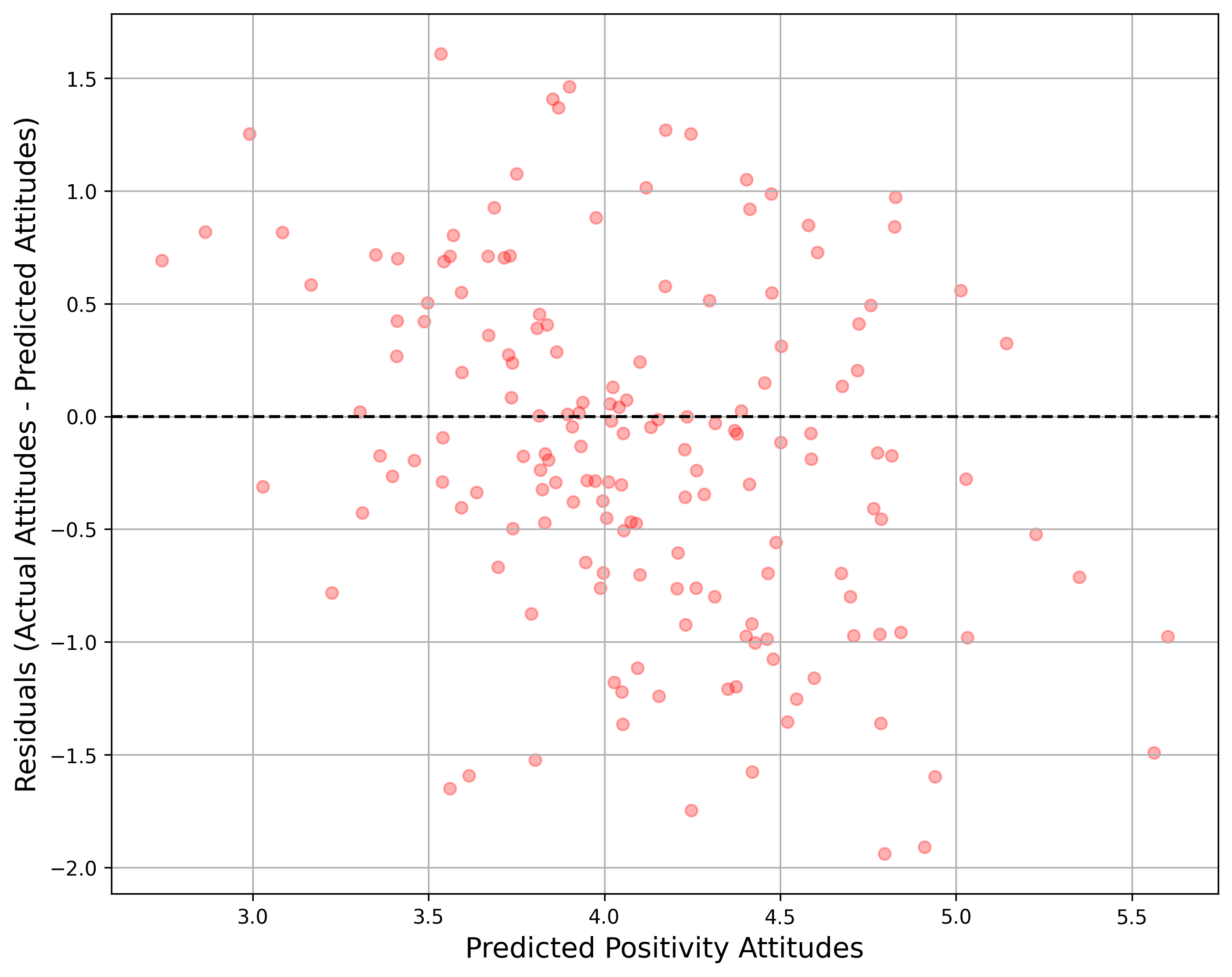}
  \end{subfigure}
\begin{minipage}{16cm}
\vspace{0.5cm}
    \footnotesize{\textit{Note:} Plots are built based on prediction and residual values for the validation set for the "Objects of Harm" outcome estimated with the best performing model - DenseNet-121 V1 (baseline model without retraining) and additional dropout, batch normalization, Dense layers, and a linear prediction layer.}
\end{minipage}
\end{figure}

\end{document}